%% file: _main.tex
\pdfoutput=1

\documentclass[11pt]{article}

\usepackage[final]{acl}

\usepackage{times}
\usepackage{latexsym}

\usepackage{booktabs}
\usepackage{amsfonts}
\usepackage{nicefrac}
\usepackage{microtype}
\usepackage{todonotes}
\usepackage{listings}
\usepackage{algorithm}
\usepackage{algpseudocode}
\usepackage{soul}
\usepackage{tabularx}
\usepackage{pifont}
\usepackage{xcolor}
\usepackage{graphicx}
\usepackage{subcaption}
\usepackage{hyperref}
\usepackage{comment}
\usepackage{tcolorbox}
\usepackage{nicematrix}
\usepackage{tabu}
\usepackage{colortbl}
\usepackage{makecell}
\usepackage{array}
\usepackage{threeparttable}
\usepackage[normalem]{ulem}

\input{macro}

\usepackage[T1]{fontenc}

\usepackage[utf8]{inputenc}

\usepackage{microtype}

\usepackage{inconsolata}

\usepackage{graphicx}

\usepackage{pifont}
\usepackage{stmaryrd}
\usepackage{hyperref}
\usepackage{multirow}
\usepackage{graphicx}
\usepackage{pdfpages}
\usepackage{url}
\usepackage{xcolor}
\usepackage{epsfig}
\usepackage{adjustbox}
\usepackage{amsfonts}
\usepackage{amsmath}
\usepackage{amssymb}
\usepackage{booktabs} 
\usepackage{comment}
\usepackage{textcomp}
\usepackage{relsize}
\usepackage{stmaryrd}
\usepackage{bbm}
\usepackage{rotating}
\usepackage{helvet}
\usepackage{natbib}
\usepackage{cleveref}
\usepackage{xspace}
\usepackage{enumitem}
\usepackage{makecell}
\PassOptionsToPackage{table}{xcolor}

\usepackage{kotex}
\usepackage{CJKutf8}

\title{Can Code-Switched Texts Activate a \textit{Knowledge Switch} \includegraphics[width=1.5em]{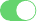} in LLMs? A~Case~Study on English-Korean Code-Switching}

\author{First Author \\
  Affiliation / Address line 1 \\
  Affiliation / Address line 2 \\
  Affiliation / Address line 3 \\
  \texttt{email@domain} \\\And
  Second Author \\
  Affiliation / Address line 1 \\
  Affiliation / Address line 2 \\
  Affiliation / Address line 3 \\
  \texttt{email@domain} \\}

\author{
    Seoyeon Kim\textsuperscript{\rm 1}~~~
    Huiseo Kim\textsuperscript{\rm 1}~~~
    Chanjun Park\textsuperscript{\rm 2}~~~
    Jinyoung Yeo\textsuperscript{\rm 1}~~~
    Dongha Lee\textsuperscript{\rm 1}\thanks{\ \ Corresponding author}\\
    \textsuperscript{\rm 1}Yonsei University~~~\textsuperscript{\rm 2}Soongsil University\\
    \texttt{\{emseoyk, pororokhs, jinyeo, donalee\}@yonsei.ac.kr}\\\texttt{chanjun.park@ssu.ac.kr}\\   
}

\begin{document}
\maketitle
\input{0_abstract}
\input{1_intro}

\input{1.5_prelim}

\input{2_dataset}

\input{3_results}

\input{5_conclusion}

\input{6_limitations}
\bibliography{custom}
\input{7_appendix}

\end{document}

%% file: macro.tex
\definecolor{mypink1}{rgb}{0.858, 0.188, 0.478}

\definecolor{olive}{rgb}{0.39, 0.875, 0.12}

\newcommand{\eg}{{\it e.g.}}%
\newcommand{\ie}{{\it i.e.}}%

\definecolor{yellow-green}{rgb}{0.3, 0.5, 0.0}

\definecolor{lightblue}{RGB}{224,236,247}
\definecolor{deepblue}{RGB}{9,46,107}
\definecolor{deepred}{RGB}{200, 0, 0}
\definecolor{defaultcolor}{gray}{0.92}
\definecolor{magenta}{RGB}{252, 48, 110}
\definecolor{teal}{RGB}{20, 150, 110}
\definecolor{lightgray}{RGB}{172, 172, 172}
\definecolor{green}{RGB}{74, 187, 83}

\newcommand{\green}{\textcolor{green}}

\newcommand{\gray}{\textcolor{gray}}
\newcommand{\lightgray}{\textcolor{lightgray}}
\newcommand{\cellgray}{\cellcolor{defaultcolor}}

\tcbset{
  aibox/.style={
    width=\textwidth,
    top=10pt,
    colback=white,
    colframe=black,
    colbacktitle=black,
    center title,
    fonttitle=\bfseries,
  }
}

\newtcolorbox{AIbox}[2][]{aibox,title=#2,#1}

\tcbset{
  aiboxsmall/.style={
    width=0.48\textwidth,
    top=10pt,
    colback=white,
    colframe=black,
    colbacktitle=black,
    center title,
    fonttitle=\bfseries,
  }
}

\newtcolorbox{AIboxSmall}[2][]{aiboxsmall,title=#2,#1}

%% file: 0_abstract.tex
\begin{abstract}

Recent large language models (LLMs) demonstrate multilingual abilities, yet they are English-centric due to dominance of English in training corpora. The limited resource for low-resource languages remains a crucial challenge. 
Code-switching (CS), a phenomenon where multilingual speakers alternate between languages in a discourse, can convey subtle cultural and linguistic nuances that can be otherwise lost in translation and elicits language-specific knowledge in human communications. 
In light of this, we investigate whether code-switching can \textit{activate}, or identify and leverage knowledge for reasoning when LLMs solve low-resource language tasks.
To facilitate the research, we first present \textsc{EnKoQA}, a synthetic English-Korean CS question-answering dataset. 
We provide comprehensive analysis on a variety of multilingual LLMs by subdividing activation process into \textit{knowledge identification} and \textit{knowledge leveraging}.
Our results demonstrate that compared to English text, CS can faithfully activate knowledge inside LLMs especially on language-specific domains, suggesting the potential of code-switching on low-resource language tasks. 

\end{abstract}

%% file: 1_intro.tex
\section{Introduction}
\input{figure_latex/motivating_ex}

Large language models (LLMs) have continuously evolved through time to exhibit advanced multilingual capabilities, enabled by training on massive datasets that include text in many different languages. However, these sources are typically skewed toward English, creating an inconsistent performance across different languages~\citep{chen-etal-2024-monolingual, zhang-etal-2024-plug}.
The limited availability for real-world user queries in low-resource languages remains a crucial challenge for achieving robust multilingual models.
Prior works attempt to mitigate this issue through machine translation~\citep{artetxe-etal-2023-revisiting, baresiss}, but crucial semantic nuances may be lost in translation, and machine translation errors are inevitable.

In human multilingual societies, code-switching (CS), or the practice of alternating between two or more languages within an utterance, is used to fill in lack of language proficiency, to emphasize certain emotions or points, or for group identity~\citep{why-code-switch}.
Moreover, code-switching functions as an effective tool to embed cultural meanings. 
Expressing certain concepts in original language can convey subtle cultural and linguistic nuances that can be lost in translation, and knowledge related to certain language are more likely to be more memorized in its own language. 
As shown in Figure~\ref{fig:motivating_ex}, when a human English-Korean bilingual is given a question that is closely related to Korean culture, a question in English and Korean code-switching is more capable of recalling knowledge about `몽유도원도'\footnote{A landscape painting by An Gyeon, commissioned by Prince Anpyeong in the early Joseon Dynasty following his dream of Shangri-la.}, because the concept is more familiar in Korean than in English. 

This observation raises intriguing insight about the impact of code-switching in multilingual societies and the potential for equivalent effect in LLMs. Given that code-switching facilitates target language-specific knowledge in human communications, we investigate whether the same applies to English-centric LLMs when solving low-resource language tasks. 
Therefore, we ask ourselves the following research question: 
\textbf{Can code-switched texts activate language-specific knowledge, or turn on a ``knowledge switch'' in LLMs?} By \textit{knowledge activation}, we refer to the overall process of identifying what knowledge is required, and applying knowledge to answer the question.

To answer the question, we subdivide knowledge activation process into two tasks: 
(1) In \textit{Knowledge Identification} task, we investigate if querying LLMs in CS and English yield different knowledge from its encoded memory. 
Specifically, we evaluate the quality of knowledge from different linguistic settings in terms of faithfulness and helpfulness.
(2) In \textit{Knowledge Leveraging} task, we observe if LLMs can faithfully ground on identified knowledge for solving question-answering (QA) task.

There have been continuous, if not abundant, researches on code-switching in the field of computational linguistics~\citep{aguilar-etal-2020-lince, rizvi-etal-2021-gcm}. 
Recently, after the emergence of LLMs with impressive multilingual abilities, a line of work have discovered LLMs' abilities in CS~\citep{huzaifah-etal-2024-evaluating, yong-etal-2023-prompting-code-mixed, zhang-etal-2023-multilingual-notyet}. However, the focus of such works are only limited to understanding and generating CS of LLMs, while the effectiveness of CS in tasks that involve low-resource language has not yet been explored. To the best of our knowledge, this work is the first to comprehensively analyze the effectiveness of code-switching on knowledge activation to LLMs.

Meanwhile, a crucial challenge when it comes to code-switching is the data scarcity. 
There is a limited number of CS datasets, let alone culture-focused data~\citep{dogruoz-etal-2021-survey}. Since CS often happens in conversations, data are not easily available and the quality is not ensured. 
To address the shortage of data, efforts have been made to synthetically generate code-switching corpus based on linguistic theories~\citep{pratapa-etal-2018-language, rizvi-etal-2021-gcm, salaam-etal-2022-offensive}. 
However, these works rely on syntactic parsers and part-of-speech taggers that support limited languages, and the quality of text are highly dependent on the performances of those tools. 
Therefore, we first construct \textsc{EnKoQA}, a synthetic English-Korean code-switching dataset to explore the potential of CS in low-resource language task.\footnote{We further discuss the resource scarcity of Korean language in Appendix~\ref{kor-low}.}
Following Matrix Language Frame Model~\citep{myers1997duelling}, we synthesize Korean QA datasets~\citep{kim-etal-2024-click, son-etal-2024-hae} that encompass various aspects of Korea into English-Korean code-switched questions. 

We conduct experiments with \textsc{EnKoQA} and provide extensive analysis on a wide range of multilingual LLMs. 
The experimental results reveal that CS is able to faithfully activate language-specific knowledge that are encoded in multilingual LLMs compared to high-resource language and target language translation; 
this tendency was more prominent on domains that specifically requires knowledge in target language and culture.

The contributions of our work are as follows:
\begin{itemize}
    \item To the best of our knowledge, this work is the first to comprehensively analyze the effectiveness of code-switching on knowledge activation to LLMs by introducing two tasks.
    \item We propose a qualified English-Korean code-switching QA dataset that is synthesized upon two Korean-centric datasets, and conduct extensive experiments on various families of multilingual LLMs. 
    \item Experimental results on extensive LLMs indicate that code-switching has advantages in knowledge activation especially on language-specific domains, suggesting the potential of code-switching text as a tool for conveying cultural nuances in target language tasks.
\end{itemize}

%% file: figure_latex/motivating_ex.tex
\begin{figure}[!t]
    \centering
    \includegraphics[width=1\columnwidth]{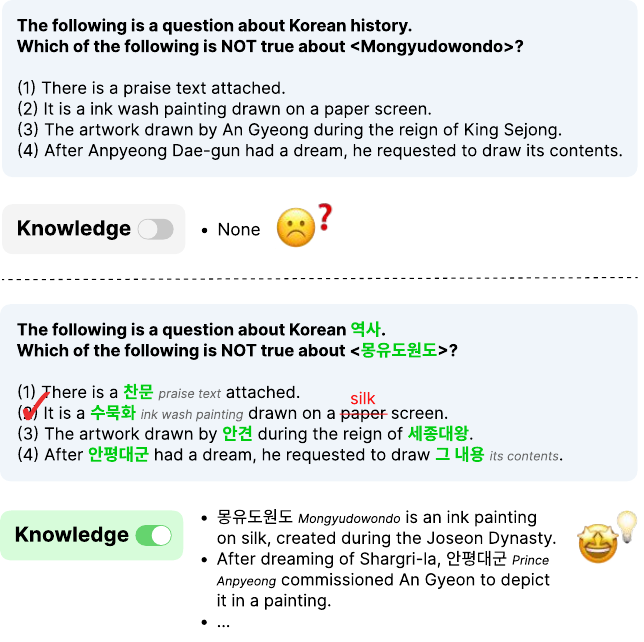}
    \caption{A motivating example of knowledge identification between languages. Compared to a question in English (\textit{top}), a bilingual speaker can ``activate'' more relevant knowledge with a question in CS (\textit{bottom}).}
    
    \label{fig:motivating_ex} 
    
\end{figure}

%% file: 1.5_prelim.tex
\section{Preliminaries \& Related Work}
In this section, we provide preliminary knowledge about code-switching, and explore relevant studies from conventional and computational linguistics. 
\subsection{Code-Switching Theories}

Many linguistic theories attempt to explain the grammatical construction of code-switched text, such as Equivalence Constraint (EC) theory and Free Morpheme Constraint (FMC) theory proposed by \citet{poplack1980}.
EC theory suggests that code-switching occurs at points in a sentence where the structures of both languages are grammatically compatible.
FMC theory suggests that code-switching cannot occur between a bound morpheme and a lexical base. (\eg{}, “He is look-ando for a book.” is a wrong code-switch.)

However, these theories have limitations in that the theory can only be applied to two language with similar or equivalent syntactic structures.
EC and FMC theories are not applicable to English-Korean code-switching text, due to the different sentence structure of Korean and English ~\citep{kor-eng-grammar}. In this regard, we adopt Matrix Language Frame Model to construct our code-switching dataset. 


\subsection{Matrix Language Frame Model}
\label{ssec:mlf}
Matrix Language Frame (MLF) model is a code-switching theory proposed by \citet{myers1997duelling}. 
MLF model posits that in any instance of code-switching, one language provides the morpho-syntactic framework of the sentence. This is known as the \textit{matrix language}. The other language, called the \textit{embedded language}, contributes to additional content, usually in the form of words or phrases, but follows the grammatical rules set by the matrix language. In other words, matrix language dominates the sentence structure, while the embedded language is integrated within that structure. Content morphemes can be in both languages, but functional morphemes come from matrix language. 
Taking Figure~\ref{fig:motivating_ex} as an example, ``그 내용'' which translates to ``its contents'' can be embedded into English sentence, but functional morpheme such as ``to'' cannot.

\subsection{Code-Switching for Language Models}
Previous works introduce benchmarks for evaluating code-switching ability of multilingual language models across multiple tasks~\citep{aguilar-etal-2020-lince, khanuja-etal-2020-gluecos}. 
More recent works focus on the capability of LLMs in code-switching. 
\citet{zhang-etal-2023-multilingual-notyet} discover performance of multilingual LLMs in various code-switching tasks, including sentiment analysis and language identification. 
\citet{yong-etal-2023-prompting-code-mixed} explore prompting multilingual LLMs to generate code-mixed data. 
\citet{shankar-etal-2024-icm} introduce a prompting technique called in-context mixing for effective in-context learning in LLMs. 
\citet{wang-etal-2025-investigating-scaling} explores the impact of code-switching on cross-lingual transfer during pre-training.
Although these benchmarks encompass a variety of tasks, the analysis of LLMs' code-switching capabilities in terms of knowledge retrieval and utilization has not yet been investigated.

\subsection{Code-Switched Data Synthesis}
Data synthesis for code-switching has been approached in various ways. 
Several studies utilize parsers and neural models to synthesize code-switched text based on EC theory~\citep{pratapa-etal-2018-language, rizvi-etal-2021-gcm}. Similarly, \citet{salaam-etal-2022-offensive} extract phrases from source language and reintegrate them into target language. 
Another line of works synthesize code-switching on token-level~\citep{li-etal-2024-prealign, wang-etal-2025-investigating-scaling}.
In recent efforts to address data scarcity in low-resource settings, LLMs have been employed to generate synthetic data~\citep{li-etal-2023-synthetic}. 
However, using LLMs specifically for synthesizing code-switched data remains unexplored.

%% file: 2_dataset.tex
\section{\textsc{EnKoQA}: English-Korean Code-Switching QA Testset}
\label{sec: dataset}
%
To compare the effectiveness of code-switching with dominant language and translation in target language when performing language-specific tasks, we introduce \textsc{EnKoQA}, a synthetic English-Korean code-switching dataset that is designed based on MLF model.
In this section, we first discuss the details of data construction (\S~\ref{ssec: construction}), and evaluate performances of LLMs on the dataset (\S~\ref{ssec: rq1 exp}, ~\ref{ssec: plain_result}).

\subsection{Dataset Construction}
\label{ssec: construction}
\paragraph{Data Sources.}
We leverage two multiple-choice Korean-centric question-answering datasets that encompass various aspects of Korean language and culture. 
CLIcK~\citep{kim-etal-2024-click} is a Korean benchmark dataset designed to test Korean cultural and linguistic knowledge collected from various official Korean exams and textbooks,~\eg{}, College Scholastic Ability Test of Korea (CSAT). 
HAE-RAE~\citep{son-etal-2024-hae} is a Korean benchmark dataset originally crafted to capture cultural and contextual nuances inherent to the Korean language, sourced from official Korean exams, textbooks, and text on the internet. 
In this work, we focus on categories about Korean society to evaluate the effect of CS on activating Korean-specific knowledge.
Specifically, we collect 1,995 pairs of eight categories from CLIcK, and 1,027 pairs of five categories from HAE-RAE, resulting in 2,372 QA pairs in nine categories: Popular, Economy, Politics, Tradition, General Knowledge, Society, Geography, History, and Law. More details of original datasets are provided in Appendix~\ref{app: datasource}.

\input{tables/main_results}

\paragraph{Automatic Translation.}
As most LLMs are trained on English-dominant corpora, we regard the English-centric LLM as a bilingual whose matrix language is English but also fairly competent in Korean. 
To generate code-switched text that follows the MLF model, we need parallel data in Korean and English to extract semantically important words or phrases from Korean text and embed into English text. 
We first automatically translate all Korean query-choices pairs into English using \texttt{gpt-3.5-turbo}, where the model is instructed to translate the query and choices to English with an one-shot demonstration. 
Lastly, human supervision was done to ensure translation quality.

\paragraph{Generating Candidates in Different Levels.}
Now that we obtain parallel data in both languages, the next step is to embed Korean content morphemes into English sentence. 
As code-switching mostly happens spontaneously, there does not exist a certain formula for mixing two languages. 
Moreover, replacing every content word with its Korean equivalent may seem rather artificial. 
To address this, we simulate a natural code-switching by creating various versions of code-switched texts at different ratios (30, 50, 70, and 90\%), then selecting a version that represents the best quality and most naturalness. 
Specifically, given a question in both languages and a specified proportion, \texttt{gpt-3.5-turbo} identifies content words from the Korean question and integrates them into the English question according to the specified proportion.
To collect contexts of various semantic importance, we employ two prompts that define ``content word'' differently; 
one defines content words as noun phrases, while the other identifies them as semantically important elements within the context.
Eight code-switched candidates are collected per question, from which human annotators select a single candidate that most faithfully follows MLF structure. 
Comprehensive details about dataset construction are provided in Appendix~\ref{app: dataset}.

\subsection{Experimental Settings}
\label{ssec: rq1 exp}
\paragraph{Models.}
We conduct extensive analysis on two groups of state-of-the-art multilingual LLMs: 
(1) Proprietary LLMs that are available via APIs, such as GPT-3.5, GPT-4o~\citep{openai2024gpt4technicalreport}, and Claude 3.5 Sonnet~\citep{claude3.5}. 
(2) Open-source LLMs such as Solar (10.7B, ~\citealp{kim2024solar107bscalinglarge}), Llama3 (8B, 70B,~\citealp{dubey2024llama3herdmodels}), and Gemma2 (9B, 27B,~\citealp{gemma2}).
More details about the models are in Appendix~\ref{app: model}.

\paragraph{Baselines.}
To compare performances of LLMs in various language settings, we evaluate on CS, English, and translated Korean (KO$_{t}$) questions. 
Korean translation baseline simulates more practical scenarios of code-switching where machine translation is adopted to convert task data from English into the target language in low-resource language tasks. 
This baseline accounts for the possibility of errors that can arise during machine translation, and to compare whether code-switching would produce more robust results in comparison.
To create KO$_{t}$, we back-translate English translation text to Korean using \texttt{gpt-3.5-turbo}.  
Prompts that we used for inference are provided in Table~\ref{tab:prompt_plain}.
We also conduct experiments on the original Korean questions, but do not consider it as major baseline, because we aim to examine the effect of code-switching compared to dominant language, rather than demonstrate the performance of low-resource language.
Please refer to Appendix~\ref{app:exp} for further discussions. 


\subsection{Results}
\label{ssec: plain_result}

\paragraph{Overall.}
As shown in Total column from Table~\ref{tab:plain_result}, the performance on CS significantly outperforms English and KO$_{t}$ across most LLMs in average. 
The gap between CS and other baselines is especially prominent in GPT-4o and Claude 3.5, where CS peaks in all domains.

\paragraph{CS questions excel at language-specific domains.}
While CS outperforms other baselines in many domains, it is worth noting that the gap between CS and English is substantially large on language-sensitive domains such as History and Tradition, both of which target language is essential for preserving information or terminology. 
Even Llama3 and Gemma2 models which relatively do not perform well on CS questions, show higher scores on CS for such domains. 
On the other hand, the phenomenon is less consistent for general knowledge (\eg, Society, General), and domains that require expert-level knowledge (\eg, Politics, Law).

\paragraph{CS surpasses translated Korean on most models.}
We compare code-switching with translated Korean translation to observe whether CS has advantages in minimizing translation errors. Except for Solar, KO$_{t}$ generally shows lowest performance among three baselines. This suggests that while translating task in target language is not the best practice, CS can faithfully encapsulate meanings and linguistic cues that may be lost in translation, highlighting the potential of leveraging CS for performing non-dominant language tasks.

\paragraph{Ratios do not affect performance.}
To ensure that the ratio of code-switching does not influence models' performances and our dataset is constructed under fair process, we calculate Code-Mixing Index (CMI) scores~\citep{srivastava-singh-2021-challenges} and report corresponding accuracy in Tradition and History domains. 
As shown in Table~\ref{tab:cmi}, we can see that accuracy scores are quite evenly distributed across all ratios, suggesting that there is no distinct tendency between CMI and accuracy.

%% file: tables/main_results.tex
\begin{table*}[!t]
\small
\centering
\resizebox{1\textwidth}{!}
{
\begin{tabular}{lc cccccccccc}
\toprule 
  Model & &
  \textbf{Economy} & \textbf{General} & \textbf{Geography} & \textbf{History} & \textbf{Law} & \textbf{Politics} & \textbf{Popular} & \textbf{Society} & \textbf{Tradition} & 
  \textbf{Total} \\
\midrule

\multirow{3}{*}{\textbf{GPT-4o}} & 
\cellgray CS & \cellgray \textbf{91.53} & \cellgray \textbf{\green{78.41}} & \cellgray \textbf{69.04} & \cellgray \textbf{\green{74.79}} & \cellgray \textbf{55.86} & \cellgray \textbf{\green{90.48}} & \cellgray \textbf{\green{95.12}} & \cellgray \textbf{63.70} & \cellgray \textbf{\green{85.14}} & \cellgray \textbf{78.23} \\
& EN & 89.83 & 75.00 & 66.19 & 61.97 & 52.64 & 84.52 & \textbf{\green{95.12}} & 60.40 & 74.32 & 73.33 \\
& KO$_{t}$ & 89.83 & 71.59 & 60.14 & 63.03 & 48.74 & 85.71 & 92.68 & 56.44 & 75.23 & 71.49 \\
\midrule

\multirow{3}{*}{\textbf{GPT-3.5}} &
\cellgray CS & \cellgray \textbf{71.19} & \cellgray 47.73 & \cellgray 44.48 & \cellgray 32.91 & \cellgray 35.40 & \cellgray \textbf{70.24} & \cellgray \textbf{80.49} & \cellgray 49.17 & \cellgray 57.21 & \cellgray \textbf{54.31} \\
& EN & \textbf{71.19} & \textbf{48.86} & \textbf{45.55} & \textbf{36.32} & \textbf{36.55} & 66.67 & 63.41 & \textbf{52.64} & \textbf{62.61} & 53.76 \\
& KO$_{t}$ & 62.71 & 26.70 & 31.67 & 26.71 & 24.83 & 48.81 & 58.54 & 37.79 & 49.55 & 40.81 \\
\midrule

\multirow{3}{*}{\textbf{Claude 3.5}} & 
\cellgray CS & \cellgray \textbf{\green{93.22}} & \cellgray \textbf{72.16} & \cellgray \textbf{\green{72.95}} & \cellgray \textbf{73.08} & \cellgray \textbf{\green{62.53}} & \cellgray \textbf{86.90} & \cellgray \textbf{\green{95.12}} & \cellgray \textbf{\green{67.66}} & \cellgray \textbf{84.23} & \cellgray \textbf{78.65}
\\
& EN & 89.83 & 71.59 & 67.97 & 61.54 & 55.63 & 85.71 & 92.68 & 63.20 & 75.23 & 73.71 \\
& KO$_{t}$ & 64.41 & 47.73 & 54.09 & 54.49 & 45.52 & 69.05 & 82.93 & 52.31 & 61.71 & 59.14 \\
\midrule

\multirow{3}{*}{\textbf{Solar}} & 
\cellgray CS & \cellgray \textbf{83.05} & \cellgray \textbf{55.11} & \cellgray 54.09 & \cellgray \textbf{63.46} & \cellgray 42.76 & \cellgray 80.95 & \cellgray \textbf{85.37} & \cellgray 54.29 & \cellgray \textbf{75.23} & \cellgray \textbf{66.03}
\\
& EN & 74.58 & 46.02 & 49.47 & 39.53 & 42.76 & 77.38 & 65.85 & 51.16 & 62.61 & 56.60 \\
& KO$_{t}$ & 81.36 & 50.57 & \textbf{56.94} & 58.12 & \textbf{46.44} & \textbf{82.14} & 78.05 & \textbf{54.95} & 70.27 & 64.31 \\
\midrule

\multirow{3}{*}{\textbf{Llama3 70B}} & 
\cellgray CS & \cellgray 79.66 & \cellgray 51.70 & \cellgray \textbf{50.53} & \cellgray \textbf{49.36} & \cellgray 44.14 & \cellgray \textbf{80.95} & \cellgray \textbf{75.61} & \cellgray \textbf{57.43} & \cellgray 65.77 & \cellgray \textbf{61.68}
\\
& EN & \textbf{83.05} & \textbf{57.39} & \textbf{50.53} & 45.94 & \textbf{45.75} & 73.81 & 73.17 & 53.30 & \textbf{66.67} & 61.07 \\
& KO$_{t}$ & 76.27 & 50.57 & 46.98 & 43.80 & 38.16 & 70.24 & 82.93 & 51.49 & 61.26 & 57.97 \\
\midrule

\multirow{3}{*}{\textbf{Llama3 8B}} &
\cellgray CS & \cellgray \textbf{69.49} & \cellgray \textbf{40.34} & \cellgray 36.30 & \cellgray 35.68 & \cellgray \textbf{35.63} & \cellgray \textbf{75.00} & \cellgray \textbf{73.17} & \cellgray 45.05 & \cellgray \textbf{54.05} & \cellgray \textbf{51.63}
\\
& EN & 64.41 & 39.77 & 37.72 & \textbf{37.39} & 32.64 & 67.86 & 63.41 & \textbf{45.21} & 53.60 & 49.11 \\
& KO$_{t}$ & 61.02 & 38.07 & \textbf{38.79} & 32.48 & 33.33 & 65.48 & 65.85 & 43.73 & 50.90 & 47.74 \\
\midrule

\multirow{3}{*}{\textbf{Gemma2 27B}} & 
\cellgray CS & \cellgray 79.66 & \cellgray 46.02 & \cellgray \textbf{48.75} & \cellgray \textbf{41.03} & \cellgray \textbf{45.29} & \cellgray \textbf{77.38} & \cellgray \textbf{78.05} & \cellgray 54.79 & \cellgray \textbf{65.32} & \cellgray 59.59
\\
& EN & \textbf{84.75} & \textbf{53.41} & 48.40 & 40.60 & 41.84 & 72.62 & \textbf{78.05} & \textbf{54.95} & 63.96 & \textbf{59.84} \\
& KO$_{t}$ & 77.97 & 44.89 & 44.84 & 41.67 & 41.84 & 73.81 & 75.61 & 50.66 & 59.46 & 56.75 \\
\midrule

\multirow{3}{*}{\textbf{Gemma2 9B}} & 
\cellgray CS & \cellgray \textbf{79.66} & \cellgray 42.05 & \cellgray 44.13 & \cellgray \textbf{40.17} & \cellgray 41.15 & \cellgray \textbf{73.81} & \cellgray 80.49 & \cellgray \textbf{53.30} & \cellgray \textbf{65.77} & \cellgray \textbf{57.84}
\\
& EN & 76.27 & \textbf{46.02} & \textbf{49.47} & 38.46 & \textbf{42.30} & 69.05 & 73.17 & 52.15 & 63.51 & 56.71 \\
& KO$_{t}$ & 76.27 & 42.05 & 41.99 & 34.62 & 40.23 & 71.43 & \textbf{82.93} & 51.98 & 58.11 & 55.51 \\
\bottomrule
\end{tabular}
}

\caption{QA performances of multilingual LLMs on CS, English, and translated Korean settings. \textbf{Bold} indicates the highest score among the three baselines from each model. \green{\textbf{Green}} indicates the highest score from each domain.}
\label{tab:plain_result}
\end{table*}

%% file: 3_results.tex
\section{Can Code-Switched Questions Activate a “Knowledge Switch” in LLMs?}
\label{sec:task}
From Section~\ref{ssec: plain_result}, we observe that most LLMs are able to answer correctly to questions in CS than in other baselines.
To further investigate on the effectiveness of CS in activating language-specific knowledge, we formulate two tasks: \textit{Knowledge Identification} and \textit{Knowledge Leveraging}. We evaluate the tasks in CS and English questions, the two baselines that share the same matrix language.


\subsection{Knowledge Identification}
\label{ssec: task_desc}



\paragraph{Task Description.}
When a human English-Korean bilingual is given a question about Korean culture, they will first try to identify what specific knowledge is required to answer the question, and then apply the knowledge to find the correct answer. 
Depending on which language the question is written in, the quantity and quality of the knowledge may vary, as described in Figure~\ref{fig:motivating_ex}. 
Language-specific knowledge is likely to be encoded much abundantly in its own language, so reading the question in CS will allow more effective knowledge activation than in English. 
In this sense, knowledge identification task evaluates LLMs' ability to identify what knowledge is prerequisite for the question. 
Specifically, the LLM is asked to write a list of factual knowledge that are necessary for solving the given question in one or two sentences. 

\paragraph{Evaluation Criteria.}
\label{sec: eval_ki}

For a qualitative analysis on knowledge identification, we evaluate the quality of a knowledge list based on two criteria: \textit{Faithfulness} evaluates whether the generated knowledge is factually correct and the model does not output hallucination.
\textit{Helpfulness} evaluates whether the knowledge is relevant to the question, and helpful for answering the question correctly. 




\subsection{Knowledge Leveraging}
\paragraph{Task Description.}
We refer to Knowledge Leveraging as applying the identified knowledge into reasoning. In specific, the model should be able to find a correct answer based on the knowledge it has identified from the Knowledge Identification task. 
Therefore, we provide knowledge identified by each model and instruct the model to find the answer using the knowledge.  
To encourage the models to properly ground on knowledge, we adopt Chain-of-Thought reasoning~\citep{cot} and prompt the models to generate reasoning steps that lead to the final answer. 
We conduct experiments on the entire dataset and report accuracy score.

\subsection{Experimental Setup}
\input{figure_latex/human_both}
 \paragraph{Implementation Details.}
We conduct experiments with the same models as in Section~\ref{ssec: rq1 exp}. 
For knowledge identification, we instruct the model to write a list of factual knowledge that are required for solving the given question in one or two sentences. 
For knowledge leveraging, we pass on previously identified knowledge and ask the model to select an answer and explain why.
The full-length prompts are provided in Table~\ref{tab:prompt_identify} and~\ref{tab:prompt_leveraging}.

\paragraph{Evaluating Knowledge Identification.}
In order to effectively evaluate knowledge identification results, we refer to Section~\ref{ssec: plain_result} and choose two domains where CS performance is higher (\textit{i.e.}, History, Tradition), and two domains that have minimum difference (\textit{i.e.}, General, Law). 
Moreover, we select four models with different performances and sizes (\textit{i.e.}, GPT-4o, Solar, Gemma2 27B, Gemma2 9B). 
Specifically, we sample 10 questions from each domain and model, resulting in 160 samples. 
Then, we conduct human and LLM-based evaluation on identified knowledge. 

\paragraph{Human Evaluation} We employ four human evaluators who are fluent in both Korean and English and completed Korean public education, thus qualified to evaluate questions sourced from Korean proficiency tests for foreigners and the Korean College Scholastic Ability Test.
For faithfulness and helpfulness, the evaluator is asked to rate a knowledge list on a Likert scale from 1 to 3. 
In pairwise evaluation, we provide two knowledge lists in a random order and ask the evaluator to select a list that is overall more effective for answering the question.
Details on evaluation criteria and evaluator information are provided in Appendix~\ref{app: humaneval} and~\ref{app:evaluator}.

\paragraph{LLM-based Evaluation} As we conduct human evaluation on quite small amount of samples, we additionally conduct LLM-as-a-judge evaluation~\citep{llm-as-judge} to amplify our analysis. Specifically, we use GPT-4o as the evaluator, using identical instructions with human evaluators on 40 questions for 9 domains and 8 models, 360 samples in total. 
Full prompts are provided in Appendix~\ref{app: humaneval}.

\input{figure_latex/human_pair}

\section{Analysis on Knowledge Identification}
\label{sec:cot_results}

\subsection{Human Evaluation}
\label{ssec: identification}


\paragraph{Faithfulness.}
In the upper row of Figure~\ref{fig:human_both}, we observe a significant gap in faithfulness scores between CS and English in both History and Tradition. 
The discrepancy is more salient in Tradition where cultural nuances is much important, implying that asking questions in CS is much successful in capturing cultural nuances and meanings. 
In General domain, the scores for CS and English are almost identical (or even better in English for Gemma2 9B), indicating that the difference in knowledge activated by CS questions compared to English questions is minimal when addressing general and common facts. 
In Law, although knowledge from CS is slightly more faithful than that from English, their absolute scores are lower than those in other domains, suggesting that models fail to identify faithful knowledge that requires domain expertise. 

\input{figure_latex/llm_pair2}

\paragraph{Helpfulness.}
The lower row of Figure~\ref{fig:human_both} presents evaluation results for helpfulness. 
It is intuitive that faithful knowledge serves as a valuable source for answering questions, and as a result, the evaluation of helpfulness shows a similar trend to that of faithfulness.
In History and Tradition, the gap between CS and English becomes larger in helpfulness, emphasizing the effectiveness of the CS setting in identifying both faithful and helpful knowledge.
It is also notable that the scores for helpfulness are particularly high for GPT-4o and Solar, models in which performance in CS surpasses that in English to a large extent (\S~\ref{ssec: plain_result}). 
In contrast, the helpfulness scores in the Law domain are considerably lower for both CS and English compared to other domains. 
Given that the Law domain requires expert-level legal knowledge, the models struggle to grasp the legal context, leading to difficulties in accurately identifying helpful knowledge sources from both CS and English questions.

\input{figure_latex/cot_radar}

\input{tables/pair_corr}

\paragraph{Pairwise Comparison.}
In Figure~\ref{fig:human_pair}, the win ratio for CS is higher in History and Tradition, demonstrating that CS questions can activate more essential knowledge sources for question answering. 
On the contrary, in domains where CS does not show its effectiveness, the win ratio of CS is comparatively lower (\textit{i.e.}, General) or the ratio of Tie is high (\textit{i.e.}, Law). Especially in the case of Law, the quality of knowledge lists generated from CS questions is evaluated as equivalent to, or even worse than, that generated from English questions.  

\subsection{LLM-based Evaluation}
We observe in Figure~\ref{fig:llm_faith} and Figure~\ref{fig:llm_help} that the score gap between CS and English in both faithfulness and helpfulness are minimal. 
In fact, CS scores are even or lower for some cases, which are inconsistent with human evaluation results. 
However, it is still worth noting that LLM-as-a-judge also assigns higher scores for advanced models, and overall scores were lower in History and Tradition. 

On the other hand, LLM judgement scores in pairwise evaluation generally agree with the human evaluations. 
We compute Cohen's Kappa ($\kappa$) score in Table~\ref{tab:pari_corr}, and follow interpretations from~\citet{landis-koch}.\footnote{\citet{landis-koch} interprets 0–0.20 as slight, 0.21–0.40 as fair, 0.41–0.60 as moderate, 0.61–0.80 as substantial, and 0.81–1 as almost perfect agreement.}
Consistent with human evaluation, the LLM judge votes CS for most cases, and the agreement is stronger with advanced models (\textit{i.e.}, GPT-4o), on culture-intensive domains (\textit{i.e.}, History, Tradition). 

While other domains fairly agree with human judgment, Law shows exceptional results.
Specifically, the LLM-as-judge evaluation reports a significantly higher win ratio for CS in the Law domain compared to human evaluation. 
However, considering that tie ratio is substantial in human evaluation as well, we speculate that LLM-as-a-judge gives a win to CS on knowledge that human evaluators regarded comparable quality with English setting.

\section{Analysis on Knowledge Leveraging}
\label{sec: leveraging}
We present the visualized results of accuracy in both CS and English settings in Figure~\ref{fig:cot_radar}, with detailed scores reported in Table~\ref{tab:cot_results}.

\paragraph{Main Observations.}
Consistent with the results in Section~\ref{ssec: plain_result}, all models demonstrate generally higher performances for CS questions compared to English questions. The results indicate that CS effectively activates knowledge across various domains while activating in dominant English language is suboptimal. GPT-4o, Claude, and Solar exhibit higher CS performance than English across all domains. 
These models not only identify faithful and helpful knowledge (\S~\ref{ssec: identification}), but also answer questions by accurately grounding on that knowledge;
this shows that CS questions robustly activate essential knowledge in these models.
On the contrary, Llama3 and Gemma2 families show poor performance in both CS and English settings in several domains, such as Geography and Law. 
Taking into account that these domains require domain-specific expertise, it is likely that their lack of understanding contributes to low accuracy, let alone CS failing to activate Korea-focused knowledge.

\paragraph{Knowledge Identification and Leveraging both matters.}
We demonstrate that qualified knowledge identification is prerequisite for knowledge activation of CS. 
The win ratio of History knowledge by GPT-3.5 was relatively poor compared to others (Figure~\ref{fig:llm_pair2}), leading GPT-3.5 to be the only model that did not benefit from CS. 
Similarly in Law, Figure~\ref{fig:human_both} and~\ref{fig:human_pair} show that helpfulness and pairwise scores for knowledge by Gemma2 9B and 27B are lower than others, which are responsible for their suboptimal performance in CS.

\paragraph{English questions hallucinate more than CS.}
Although we informed the models that the answer is in one of the choices, we notice that the majority of incorrect responses were ``None of the above''. The errors may derive from either hallucinated knowledge or failing to follow instructions faithfully. Therefore, we provide additional analysis on erroneous outputs in Table~\ref{tab:none}. We report the results in the format of  \textit{\# of errors that derived from knowledge hallucination} / \textit{total \# of None errors}. Errors that are not from hallucination are caused by poor instruction-following.
Overall, we observe that answering English questions results in more errors compared to CS across all LLMs, and most of them were hallucinations. 
This indicates that models hallucinate much frequently when English questions are given, again highlighting the effectiveness of CS over English. 
It is also worth noting that Gemma2 families hallucinate largely on History and General, supporting our finding in Figure~\ref{fig:human_both} and~\ref{fig:cot_radar} which respectively illustrates poor performance on human evaluation and QA accuracy.

\paragraph{Case Study.}
We examine a sample case to compare the capability of code-switching and English on knowledge activation. Table~\ref{tab:case_study1} shows the knowledge and answer generated by Solar in Tradition. 
The question asks about `정월대보름 \textit{(Jeong-wol Dae-bo-reum)}', a Korean traditional holiday that celebrates the first full moon of lunar new year.
We observe that CS question preserves unique terms such as `정월대보름' and `귀밝이술 \textit{(Gwi-bal-ki-sul)}' in Korean;
this helps the model to successfully activate faithful knowledge, consequently leading to the correct answer. 
However, in the case of English, not only are these cultural nuances lost in English question, but the model misunderstood the question to asking about `단오 \textit{(Dan-o)}', another Korean traditional holiday.
Solar lacks in knowledge about `정월대보름' in English, or fails to activate encoded knowledge with its English translation.

We also provide a case of CS failing in knowledge activation in Table~\ref{tab:case_study2}.
In the case of Gemma2 9B on Law domain, hallucinations are observed in the knowledge generated from CS question. According to the Civil Act of the Republic of Korea, individuals under the age of 14 can only enter into binding contracts with the consent of their legal guardians. Additionally, individuals between the ages of 14 and 19 are not deprived of contractual effect; rather, they are granted the right to cancel such agreements at their discretion.
Moreover, the knowledge generated in English incorrectly applies the U.S. standard, which defines minors as those under 18 years of age, instead of the Korean standard, which applies to individuals under 19 years of age. This finding suggests that English question is not helpful for identifying necessary and language-specific knowledge.

%% file: figure_latex/human_both.tex
\begin{figure}[!t]
    \centering
    \includegraphics[width=1\columnwidth]{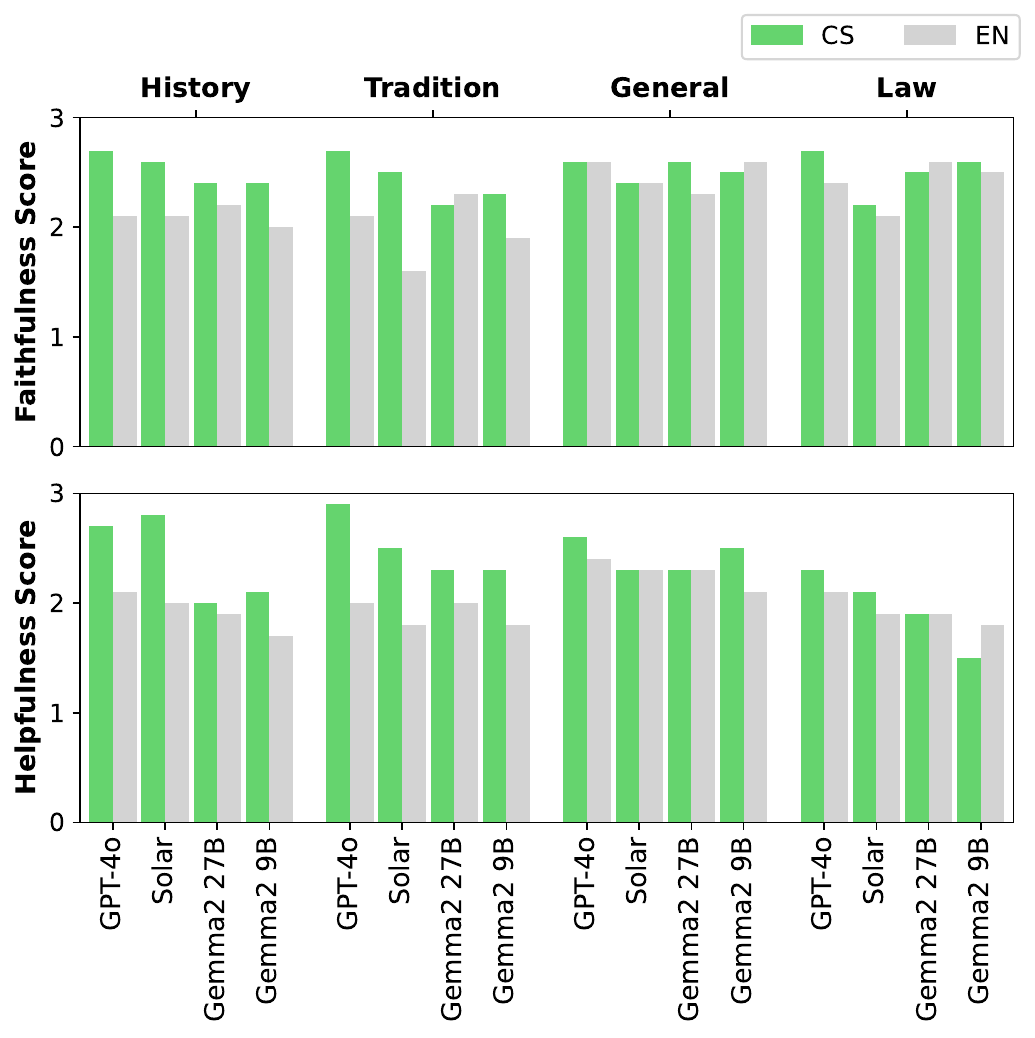}
    
    \caption{Human evaluation results on faithfulness (\textit{top}) and helpfulness (\textit{bottom}) of knowledge lists  identified from \green{\textbf{CS}} questions and \gray{\textbf{English}} questions.}
    
    \label{fig:human_both} 
    
\end{figure} 

%% file: figure_latex/human_pair.tex
\begin{figure}[!t]
    \centering
    \includegraphics[width=1\columnwidth]{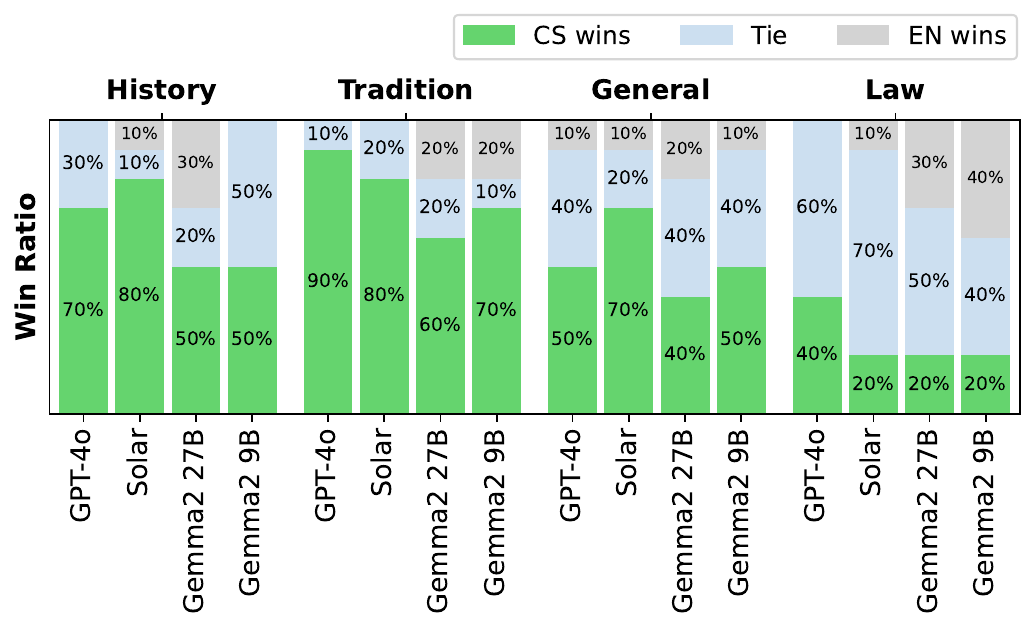}
    
    \caption{Human evaluation results on pairwise comparisons between knowledge lists identified from \green{\textbf{CS}} questions and \gray{\textbf{English}} questions.}
    
    \label{fig:human_pair} 
    
\end{figure}

%% file: figure_latex/llm_pair2.tex
\begin{figure}[!t]
    \centering
    \includegraphics[width=0.5\textwidth]{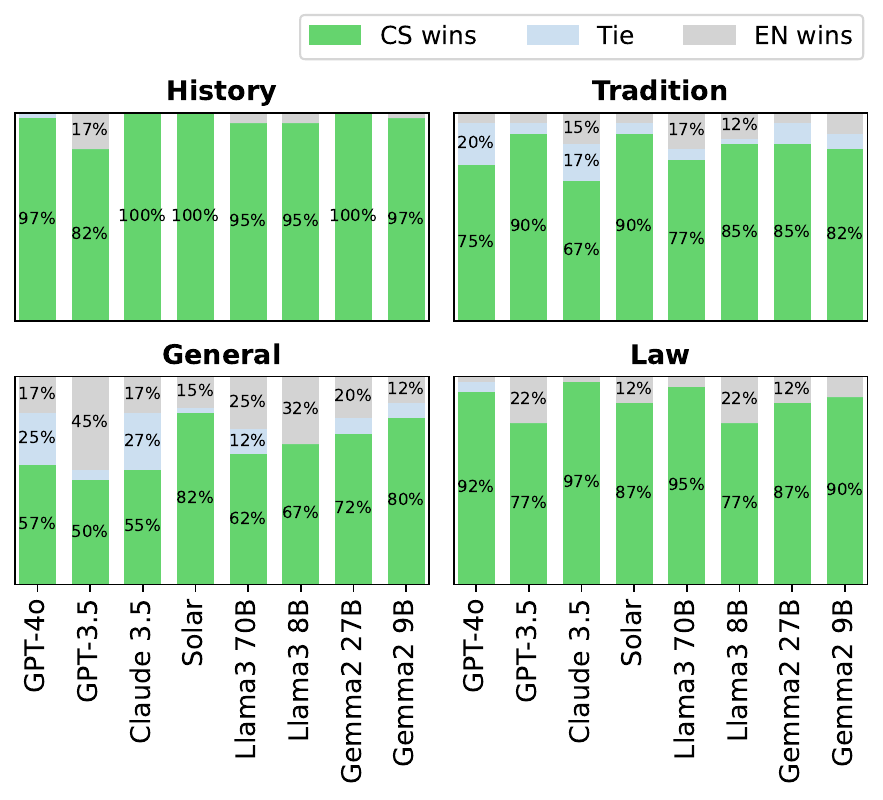}
    
    \caption{LLM-as-a-judge evaluation results on pairwise comparison between knowledge lists identified from \green{\textbf{CS}} questions and \gray{\textbf{English}} questions.}
    
    \label{fig:llm_pair2} 
    
\end{figure}

%% file: figure_latex/cot_radar.tex
\begin{figure*}[t]
    \centering
    \includegraphics[width=1\textwidth]{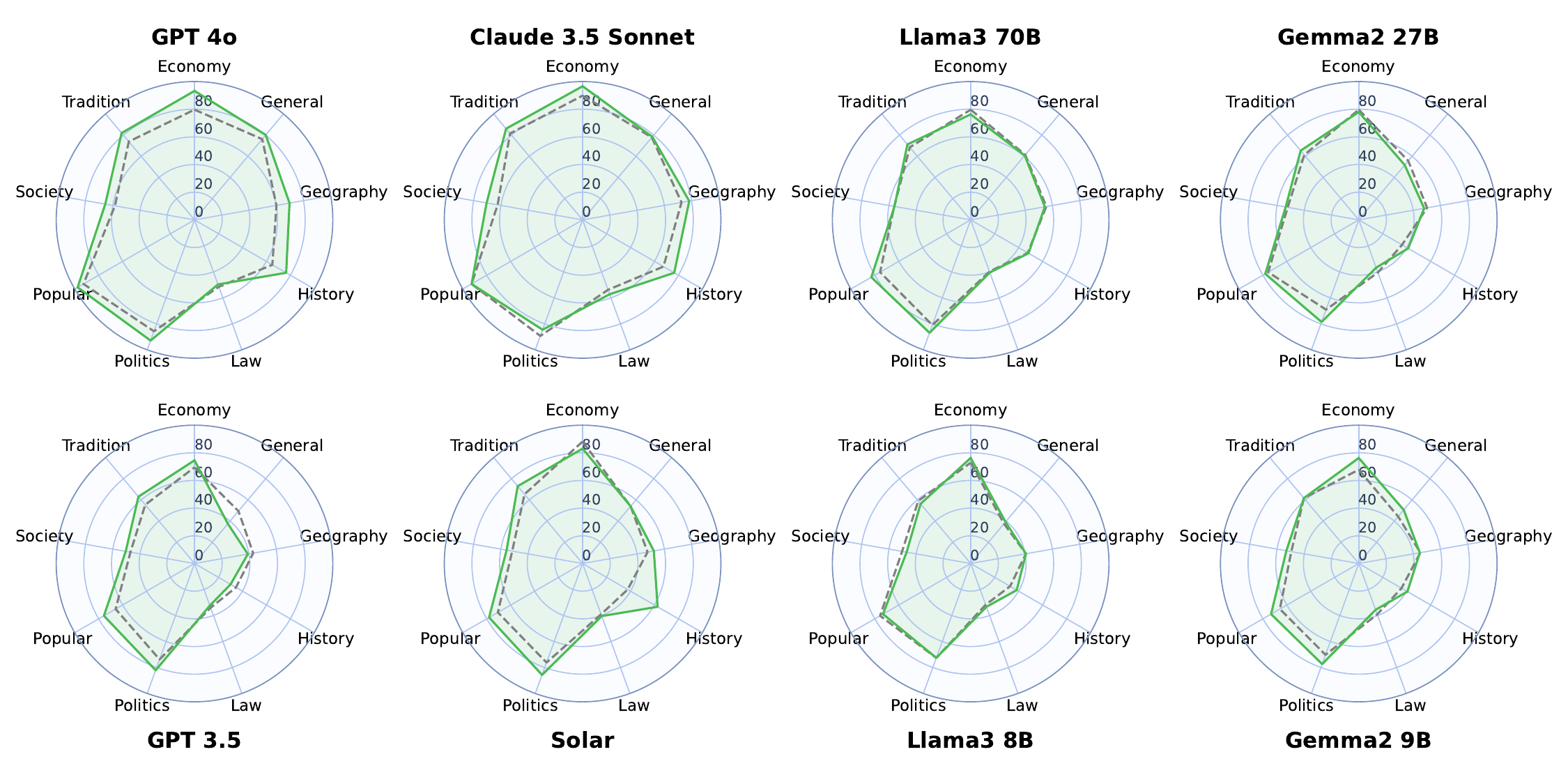}
    
    \caption{Radar charts of knowledge leveraging performances on all domains across various multilingual LLMs. \green{\textbf{Green}} line is code-switching and dashed \gray{\textbf{gray}} line is English. We report accuracy for the evaluation metric.
    }
    
    \label{fig:cot_radar} 
    
\end{figure*}

%% file: tables/pair_corr.tex
\begin{table}[!t]
\small
\resizebox{1\columnwidth}{!}
{
\begin{tabular}{lcccc}
\toprule 
  Model &
  \textbf{History} & \textbf{Tradition} & \textbf{General} & \textbf{Law} \\
  \midrule
  \textbf{GPT-4o} & 0.41 & 0.64 & 0.62 & 0.62 \\
  \textbf{Solar} & 0.26 & \lightgray{-0.09} & 0.38 & \lightgray{0.02} \\
  \textbf{Gemma2 27B} & 0.25 & 0.52 & \lightgray{0.17} & 0.34 \\
  \textbf{Gemma2 9B} & 0.20 & \lightgray{-0.07} & \lightgray{0.05} & 0.24 \\
\bottomrule

  \end{tabular}
}

\caption{Cohen's kappa ($\kappa$) correlation scores between human and LLM-as-a-judge evaluation. \lightgray{Gray} indicates poor agreement.}
\label{tab:pari_corr}
\end{table}

%% file: 5_conclusion.tex
\section{Conclusion and Future Work}
We explore the efficacy of code-switching in activating language-specific knowledge embedded in LLMs. 
Utilizing two Korean-centric QA datasets, we synthesize \textsc{EnKoQA}, a qualified English-Korean code-switching QA dataset and conduct experiments on various multilingual LLMs. 
Our analyses demonstrate that LLMs can simulate a similar code-switching effect with human communications of facilitating low-resource knowledge within LLMs, particularly in language-specific domains. 
Regarding this finding, we suggest that code-switching can be an effective strategy for solving low-resource language tasks. Also, augmenting low-resource datasets into code-switching text can amplify resource and mitigate data scarcity challenge.

Future directions of this work involve a more comprehensive investigation into the potential of code-switching across different aspects of multilingual modeling, including its role in pretraining and instruction-tuning, as well extending our work to a wider range of languages. By first addressing the relatively understudied Korean-English pair, our research contributes to a relatively unexplored area and serves as a stepping stone for subsequent studies on code-switching. In particular, our data synthesis approach, grounded in the universally applicable MLF theory, extends beyond Korean and is generalizable to all languages. Through our work, we encourage the NLP community to further explore the capacity of code-switching to effectively enhance and harness the potential of multilingual LLMs.

%% file: 6_limitations.tex
\section*{Limitations}
In this work, we focus on code-switching between English and Korean, specifically limiting the scope to Korea-specific knowledge. 
However, it is important to note that this study serves as a starting point focused on the Korean context and leaves room for expanding the scope of code-switching to other cultures and languages. 
For future research, we aim to investigate whether the knowledge activation effect also occurs in other language settings.

Another limitation of our work is that we conduct human evaluations on only a subset of LLMs, domains, and questions. Evaluating the quality (\textit{i.e.}, faithfulness and helpfulness) of knowledge in code-switched text presents inherent and practical challenges, as it necessitates evaluators to be fluent bilinguals. 
Consequently, we present only partial results for the knowledge identification task. 


Lastly, as we rely on a LLM, specifically \texttt{gpt-3.5-turbo}, to synthesize our code-switching dataset, the performance of the LLM can affect the quality of the dataset. 
To mitigate the risk of erroneous samples and to fully leverage the LLM's capabilities, we engage reliable human annotators to review the samples and verify their quality. Also, as we formulate our code-switching dataset with gold English and Korean, in a more realistic scenario where a monolingual English speaker creates code-switching text, sentences would be created automatically without any additional supervision.





\section*{Ethical Consideration}
Our work utilizes large language models for data construction. Recent work has highlighted the risks of LLMs in hallucination~\citep{zhang2023siren}. In order to prevent any hallucination or harmful contents, we ensure that human annotators examined each sample carefully and create dataset safely.

\section*{Acknowledgements}
This work was supported by the IITP grants funded by the Korea government (MSIT) (No. RS-2020-II201361; RS-2024-00457882, AI Research Hub Project), and the KBSI (National research Facilities and Equipment Center) grant funded by the Korea government (MSIT) (No. RS-2024-00403860).

%% file: 7_appendix.tex
\appendix
\section{Dataset Details}
\label{app: dataset}

\subsection{Details of Source Data}
\label{app: datasource}
CLIcK~\citep{kim-etal-2024-click}\footnote{\url{https://huggingface.co/datasets/EunsuKim/CLIcK}} consists of 1,995 multiple-choice QA pairs, classified in two main categories (Culture, Language) and 11 sub-categories. 
CLIcK is sourced from various official Korean exams and textbooks,~\eg{}, College Scholastic Ability Test of Korea (CSAT). 
In this work, we only utilize data of eight sub-categories from Korean Culture category as our work aims to evaluate the effect of CS on activating Korean-specific knowledge.

HAE-RAE~\citep{son-etal-2024-hae}\footnote{\url{https://huggingface.co/datasets/HAERAE-HUB/HAE_RAE_BENCH_1.1}} is a Korean benchmark dataset originally crafted to capture cultural and contextual nuances inherent to the Korean language. We use 1,027 multiple-choice QA pairs regarding Korean culture.
Both datasets are sourced from official Korean exams, textbooks, and text on the internet. 

We combine two datasets and merge common categories (\ie{}, Society, Geography, and Law), resulting in 2,372 QA pairs in nine categories: Popular, Economy, Politics, Tradition, General Knowledge, Society, Geography, History, and Law.

\subsection{Dataset Statistics and License}
\label{app: datastat}
We provide statistics of EnKoQA per domain in Table~\ref{tab:dataset_stat}.
We plan to release the dataset in public, under CC BY-NC license. We clarify that the source datasets are either open-source or used under authors' permission, ensuring that there are no issues regarding their use.
\input{tables/dataset_stat}

\subsection{Code-Mixing Index}
\label{app: cmi}
\input{tables/cmi}

We report CMI scores for our dataset in Table~\ref{tab:cmi}. In specific, we tokenized the sentence using \texttt{bert-base-multilingual-cased}, then removed all noisy tokens such as numbers or tags and counted the ratio of 
\begin{math}\frac{num\ of\  Korean\ tokens}{num\ of\ all\ tokens}\end{math}.
We report the distribution of QA accuracy on different CMI scores in Tradition and History, two domains where CS proved its effectiveness. If CMI is close to 0, sentence is mostly written in English, and close to 100 means vice versa. The number of samples at each end (0-10, 90-100) was very small, causing outliers. We can see that accuracy scores are quite evenly distributed across all ratios, suggesting that there is no distinct tendency between CMI and accuracy performance.

It is important to note, however, that code switching metrics such as CMI, while offering a quantitative measure of token-level composition, are inherently limited in capturing the nuanced semantic and syntactic characteristics of code-switched texts. These metrics primarily rely on surface-level token ratios, which can inadvertently assign high scores to linguistically or contextually meaningless sequences. Consequently, they may over-represent the presence of meaningful code-switching patterns while failing to account for the deeper linguistic interplay that defines effective code-switching. For a more comprehensive discussion of these limitations, please refer to \citealp{srivastava-singh-2021-challenges}.

\subsection{Quality Control Guideline}
\label{app: qcguideline}
We provide a guideline we used to filter the candidates and select the final candidate.
\begin{itemize}
    \item Is the question written in English-Korean code-switching, where matrix language is English and semantically important Korean words are embedded into English sentence?
    \item Do choices also follow the code-switched pattern of query?
    \item Does the syntactic structure of the sentence follow that of English?
\item Are semantically important nouns and noun phrases from Korean sentence, and are they embedded into English sentence?
\item Are functional words and grammatical morphemes kept in English?
\end{itemize}

\subsection{Annotation Details}
For dataset construction, two Korean native annotators with expert knowledge in Korean culture and equivalently fluent in English manually examine the candidates and select the most naturally code-switched question, then cross-checked each other’s assigned share of dataset. If a selected candidate appeared to be incorrect or suboptimal, the annotators engaged in thorough discussions until they reached an agreement on the most appropriate candidate.

Regarding inter-annotator agreement (IAA), although we did not compute a formal IAA score, significant effort was devoted to ensuring high annotation quality through extensive discussion and collaboration among annotators. 
The task for human annotators in our datasets was not labeling answers, but verifying the quality of the dataset. IAA scores are for quantifying annotator agreement and further improving the dataset. In that sense, our qualitative human verification comparably guarantees both annotator consensus and dataset quality. We refined the dataset until it reached perfect agreement between the annotators, which is comparable to 1 in IAA score.
In specific, the annotators who are fluent in both English and Korean are assigned each portion of the dataset to select a candidate for code-switched question.
Following this initial annotation, the annotators cross-checked each other’s work to identify any discrepancies. If a selected candidate appeared to be incorrect or suboptimal, the annotators engaged in thorough discussions until they reached an agreement on the most appropriate candidate. This iterative and collaborative process was integral to constructing a high-quality dataset.

\subsection{Dataset Size and Quality}
\paragraph{Discussion on Dataset Size} While we acknowledge the relatively limited size of EnKoQA dataset, we emphasize that quality often matters more than quantity as many studies~\citep{pacchiardi2024100instancesneedpredicting, pmlr-v235-maia-polo24a, vivek-etal-2024-anchor} have demonstrated. Please note that we prioritized creating a high-quality dataset with rigorous manual validation and linguistic alignment, ensuring that the dataset serves as a reliable resource for code-switching research.
Additionally, while the size of Korean datasets is often limited given that Korean is a low-resource language, EnKoQA dataset is comparatively larger than the sizes of other Korean datasets. For instance, datasets in the Open Ko-LLM leaderboard~\citep{open-ko-llm}, such as Ko-ARC (1.1k), Ko-TruthfulQA (0.8k), and Ko-CommonGen (0.8k), are all smaller in scale than EnKoQA’s 2,372 question-answer pairs. This highlights our effort to provide a relatively extensive resource within the constraints of dataset availability for minor languages.

Specifically, our quality control process includes human annotators thoroughly reviewing all LLM-generated samples to assess the quality and naturalness. When any errors or unnatural code-switching patterns were identified, annotators corrected them to ensure that the final dataset adheres to high standards of our quality control. In that sense, GPT-3.5-turbo served as an assistive tool for providing initial candidates, rather than generating final outputs. Therefore, we assert that any potential shortcomings of the translation tool were effectively mitigated through this meticulous human review and correction process.

\input{tables/data_sample}

\paragraph{Translating with GPT-3.5} We have conducted experiments on both GPT-3.5 and GPT-4o for translation and code-switching generation tasks. Interestingly, we observed that after manual examination and correction process, the results from both models were comparable in terms of quality and naturalness. This is due to our rigorous human-in-the-loop workflow that ensures any errors or unnatural expressions are taken care of, regardless of the initial model used.
Given this finding, we used GPT-3.5 for its cost efficiency while maintaining high-quality standards through meticulous human examination and refinement. By prioritizing manual validation, we ensured that the final dataset reflects linguistic accuracy and naturalness, independent of the model used for preliminary generation.

\subsection{Data Sample}
We also provide a sample of original Korean, translated English, and synthesized CS example question in Table~\ref{tab:data_sample}. Note that unique terms or semantically important words are properly embedded in Korean.

\subsection{Resource Scarcity of Korean Language}
\label{kor-low}
There are varying perspectives on whether or not Korean is a \textit{low-resource} language. The bottom line is, there is no clear consensus on the term `low-resource' itself. Low-resource and high-resource are generally distinguished based on the availability of data resources and the size of the language’s speaker population~\citep{nigatu-etal-2024-zenos}. Some papers have classified Korean as low-resource~\citep{lee-etal-2024-commit}, while others have described it as mid-resource~\citep{joshi-etal-2020-state}.
What is certain is that Korean has fewer resources compared to high-resource languages (e.g., English, Chinese, Spanish, or Arabic), and particularly in the field of code-switching, there is only a handful of research or data. In our study, we aimed to compare cases where the LLM operates in the dominant language with cases where code-switching involving the target language is used. It is in this context that we referred to Korean as a low-resource language.

\section{Experimental Details}
\subsection{Computational Resources and API Cost.}
\paragraph{Llama3 and Gemma2 models.}
We used Huggingface model cards and run them on two NVIDIA A100 GPUs. Specifically, we used \texttt{meta-llama/meta-llama-3-8b-instruct}, \texttt{meta-llama/meta-llama-3-70b-instruct}, \texttt{google/gemma-2-9b-it}, \texttt{google/gemma-2-27b-it}.

\paragraph{GPT-3.5 and GPT-4o.}
We used up-to-date versions of \texttt{gpt-3-5-turbo} and \texttt{gpt-4o} APIs. The cost for \texttt{gpt-3-5-turbo} was \$15 for EnKoQA generation and \$6 for experiment inference, while the cost for \texttt{gpt-4o} was \$23 for experiment inference.

\paragraph{Claude 3.5.}
We used \texttt{claude-3-5-sonnet} API from Anthropic AI\footnote{\url{https://www.anthropic.com/}}. The cost for \texttt{claude-3-5-sonnet} was \$21 for experiment inference.

\paragraph{Solar}
We used \texttt{solar-mini} API from Upstage\footnote{\url{https://www.upstage.ai/}}. 
\label{app: model}

\subsection{Prompts}
We provide the following prompts used in our experiments. Table~\ref{tab:prompt_dataset} contains the prompt used for generating code-switched text candidates across different levels of linguistic complexity. For QA inference tasks, we used the prompt presented in Table~\ref{tab:prompt_plain}. The prompt for identifying relevant knowledge in a given context is provided in Table~\ref{tab:prompt_identify}, while Table~\ref{tab:prompt_leveraging} shows the prompt used for leveraging this identified knowledge in downstream tasks.
\label{app:prompt}

\subsection{Comparing CS and original Korean}
\label{app:exp}
We also present experimental results on original Korean data in Table~\ref{tab:og_kor}. 
Generally, performances in original Korean are higher than in CS, while CS score approximates or equal to in many cases. Considering the English dominance of the LLMs, GPT-4o and Claude 3.5 Sonnet present advanced multilingual ability with over 80\% accuracy. 
As our main research focus was towards on examining the effect of code-switching compared to dominant language, we exclude original Korean as our major concern.
Also, there is a severe possibility that the performances may be influenced by already having seen the datasets, \ie{}, data contamination, as the original datasets are sourced from official exams and texts from the Internet that are openly available.

\subsection{Open-ended QA.}
Out dataset, \textsc{EnKoQA} is multiple-choice QA dataset, following its original source datasets. We additionally explore the potential of code-switching on open-ended QA as well. 

Results are shown in Table~\ref{tab:openqa}. Using same questions in our dataset, we instruct the model to respond in short answer and compute exact match score. It is noticeable that the performances are very low compared to multiple-choice QA results. We attribute this to the free-form response of open-ended tasks, causing more errors and hallucinations. It is observable that the models barely answer correctly in History and Popular. 
\input{tables/openqa}

\section{Evaluation Details}
\subsection{Evaluation Criteria}
\label{app: humaneval}
We provide evaluation guideline for human evaluation. 

\paragraph{Faithfulness.}
Faithfulness evaluates the factual correctness of the knowledge. 
\begin{itemize}
    \item Knowledge list is very faithful. Every knowledge is factually correct.
    \item Knowledge list is somewhat faithful. Some, not every, knowledge is factually correct.
    \item Knowledge list is not faithful at all. Every knowledge is hallucinated.
\end{itemize}

\paragraph{Helpfulness.}
Helpfulness evaluates how useful the knowledge is for answering the question.
\begin{itemize}
    \item Knowledge list is very helpful. Every knowledge is relevant to the question, and used for finding the answer.
    \item Knowledge list is somewhat helpful. Some, not every, knowledge is useful for finding the answer.
    \item Knowledge list is not helpful at all. All knowledge are irrelevant with the question.
\end{itemize}

\paragraph{Pair-wise comparison.}
We comprehensively evaluate the quality of knowledge generated from CS and English questions in terms of both faithfulness and helpfulness. If both are identical, evaluators can choose \textit{Tie}.

In case of LLM-as-a-judge evaluation, same criteria and instructions are given as prompts. 

\subsection{Human Evaluator Qualifications}
\label{app:evaluator}
For knowledge identification evaluation, collecting qualified bilingual evaluators was not easy due to the inherent challenge in code-switching research of necessitating fluent bilinguals as evaluators. 
We managed to collect four Korean graduate school students as our evaluators, all of whom are native Korean with sufficient understanding of Korean culture. Also, they possess qualified English exam scores, indicating that they have no problem in understanding Korean-English code-switched texts. To mitigate the shortage of labor force, we designed the evaluation criteria objectively, allowing for an assessment that is not subjective and has clear correct answers. Specifically, we evaluate knowledge identification based on two criteria: faithfulness and helpfulness. Faithfulness evaluates the factualness of the knowledge, so the evaluators are required to use their background knowledge as well as searching from faithful sources where gold knowledge exists. To evaluate helpfulness, evaluators are given a gold answer to the question and determine whether the knowledge is helpful for finding the answer, using their logical reasoning.

\section{Observations}
In this section, we provide additional results and comprehensive observations throughout our work.

\subsection{Knowledge Identification Results}
We observed that the majority of models benefitted from CS questions. Table~\ref{tab:plain_result} shows that scores in CS are higher on all models in Politics, and in case of Law, only three models (GPT-3.5, Llama3 70B, and Gemma2 9B) out of eight models performed worse.
We can see in Average score, all models except Gemma2 27B performed better on CS.

\subsection{Knowledge Leveraging Results}
\input{tables/cot_results}

We provide accuracy results of Knowledge Leveraging in Table~\ref{tab:cot_results}. Figure~\ref{fig:cot_radar} is a visualization of this table.

\subsection{Error Analysis}
\input{tables/none}

We provide full results of error counts in Table~\ref{tab:none}. Note that as models get smaller and show poor performance in Korean, the number of errors increase. (See Gemma2 families.)

\input{figure_latex/llm_faith}
\input{figure_latex/llm_help}
\input{figure_latex/llm_pair}
\input{tables/og_kor_result}

\input{tables/case_study1}

\input{tables/case_study2}

\input{tables/prompt_dataset}

\input{tables/prompt_plain}
\input{tables/prompt_identify}

\input{tables/prompt_leveraging}

%% file: tables/dataset_stat.tex
\begin{table}[h]
\centering
{
\begin{tabular}{lr}
\toprule 
  Domain & \#
  \\
\midrule
Economy & 59   \\
General & 176   \\
Geography & 281 \\
History & 468   \\
Law & 435       \\
Politics & 84  \\
Popular & 41   \\
Society & 606   \\
Tradition & 222 \\
\midrule
Total & 2,372 \\

\bottomrule

\end{tabular}
}

\caption{Number of samples in EnKoQA.}

\label{tab:dataset_stat}
\end{table}

%% file: tables/cmi.tex
\begin{table*}[h]
    \centering
\resizebox{1\textwidth}{!}
{
    \begin{tabular}{c|cccc|cccc}
    \toprule
    \textbf{CMI} & \multicolumn{4}{c|}{\textbf{Tradition}} & \multicolumn{4}{c}{\textbf{History}} \\
    \midrule
    & \textbf{Solar} & \textbf{Gemma 2 9B} & \textbf{Gemma 2 27B} & \textbf{GPT-4o} & \textbf{Solar} & \textbf{Gemma2 9B} & \textbf{Gemma2 27B} & \textbf{GPT-4o} \\
    \midrule
    0--10 & 00.00 & 00.00 & 00.00 & 00.00 & 00.00 & 00.00 & 00.00 & 00.00 \\
    \midrule
    10--20 & 100.0 & 100.0 & 100.0 & 100.0 & 50.00 & 50.00 & 100.0 & 50.00 \\
    \midrule
    20--30 & 56.25 & 65.62 & 53.12 & 71.88 & 50.00 & 31.58 & 36.84 & 60.53 \\
    \midrule
    30--40 & 72.34 & 59.57 & 55.32 & 76.6 & 66.67 & 48.72 & 45.3 & 77.78 \\
    \midrule
    40--50 & 80.39 & 62.75 & 70.59 & 84.31 & 69.54 & 42.38 & 42.38 & 78.15 \\
    \midrule
    50--60 & 83.72 & 74.42 & 74.42 & 93.02 & 62.35 & 40.00 & 35.29 & 74.12 \\
    \midrule
    60--70 & 85.19 & 66.67 & 66.67 & 92.59 & 50.00 & 28.85 & 46.15 & 63.46 \\
    \midrule
    70--80 & 66.67 & 58.33 & 66.67 & 91.67 & 57.89 & 15.79 & 21.05 & 57.89 \\
    \midrule
    80--90 & 66.67 & 83.33 & 83.33 & 100.0 & 50.00 & 50.00 & 25.00 & 100.0 \\
    \midrule
    90--100 & 00.00 & 00.00 & 00.00 & 00.00 & 00.00 & 00.00 & 00.00 & 00.00 \\
    \bottomrule
    \end{tabular}
}

    \caption{Distribution of QA accuracy on different CMI scores in Tradition and History. If CMI is close to 0, sentence is mostly written in English, and close to 100 means vice versa. The number of samples at each end (0-10, 90-100) was very small, causing outliers.}
    \label{tab:cmi}
\end{table*}

%% file: tables/data_sample.tex
\begin{table*}[h]
\small
    \centering
    \resizebox{0.99\textwidth}{!}{%
    \begin{tabular}{cll}
        \toprule
        \textbf{Lang} 
        & \textbf{QUESTION}
        & \textbf{CHOICES}

        \\\midrule

        \makecell*[l]{KO\\}
        &
        \makecell*[{{p{7cm}}}]{
        다음 글의 (가)에 대한 (나)의 상대적 특성으로 옳은 것은? (단, (가), (나)는 각각 겨울과 여름 중 하나임.)\\
        우리나라는 더위와 추위에 대비하여 대청마루와 온돌 같은 전통 가옥 시설이 발달하였다. 대청마루는 바람을 잘 통하게 하여 (가) 을 시원하게 지낼 수 있도록 설치되었다. 온돌은 아궁이의 열을 방으로 전달하여 (나) 을 따뜻하게 지낼 수 있도록 설치되었다. 대청마루 는 중부와 남부 지역에 발달한 한편, 온돌은 대부분의 지역에 발달하였다.\\
        }
        &
        \makecell*[{{p{6.5cm}}}]{
        (1) 평균 상대 습도가 높다.\\
        (2) 정오의 태양 고도가 높다.\\
        (3) 한파의 발생 일수가 많다.\\
        (4) 대류성 강수가 자주 발생한다.\\
        (5) 열대 저기압의 통과 횟수가 많다.\\
        } 
        \\\midrule

        \makecell*[l]{EN\\}
        &
        \makecell*[{{p{7cm}}}]{
        What is the correct relative characteristic of (나) in relation to (가) in the following passage? (Note that (가) and (나) refer to either winter or summer.)\\
        In Korea, traditional house facilities such as daecheongmaru and ondol have developed to cope with heat and cold. Daecheongmaru is designed to allow good ventilation to keep (가) cool. Ondol transfers heat from the kitchen stove to the room to keep (나) warm. While daecheongmaru is developed in the central and southern regions, ondol is developed in most areas.\\
        }
        &
        \makecell*[{{p{6.5cm}}}]{
        (1) The average relative humidity is high.\\
        (2) The midday sun's altitude is high.\\
        (3) There are many days of occurrence of cold waves.\\
        (4) Heavy rainfall often occurs in Daeryuseong.\\
        (5) There are many occurrences of passage of tropical cyclones.\\
        } \\\midrule

        \makecell*[l]{CS\\}
        &
        \makecell*[{{p{7cm}}}]{
        What is the correct relative characteristic of (나) in relation to (가) in the following passage? (Note that (가) and (나) refer to either winter or summer.)\\
        In 한국, 전통 가옥 시설 such as 대청마루 and 온돌 have developed to cope with heat and cold. 대청마루 is designed to allow good ventilation to keep (가) cool. 온돌 transfers heat from the kitchen stove to the room to keep (나) warm. While 대청마루 is developed in the 중부 and 남부 지역, 온돌 is developed in most areas.
        }
        &
        \makecell*[{{p{6.5cm}}}]{
        (1) The average 상대 습도 is high.\\
        (2) The 정오의 태양 고도 is high.\\
        (3) There are many days of occurrence of 한파.\\
        (4) 대류성 강수 often occurs.\\
        (5) There are many occurrences of passage of 열대 저기압.\\
        }
        \\\bottomrule
        
    \end{tabular}}
    \caption{An example of Korean, English, and CS from dataset.}
    \label{tab:data_sample}
\end{table*}

%% file: tables/openqa.tex
\begin{table*}[ht]
     \small
    \centering 
\resizebox{1\textwidth}{!}
{
\begin{tabular}{lcccccccccccccccccccccccccccc}
\toprule
\textbf{Model} &&\textbf{Economy} & \textbf{Geography} & \textbf{History} & \textbf{Law} & \textbf{Politics} & \textbf{Popular} & \textbf{Society} & \textbf{Tradition} \\
\midrule
\multirow{3}{*}{\textbf{GPT-4o}} 
& CS & 85.00 & 20.00 & 40.00 & 30.00 & 30.00 & 05.00 & 50.00 & 35.00 \\
& EN & 80.00 & 00.00 & 05.00 & 05.00 & 10.00 & 00.00 & 05.00 & 00.00 \\
& KOR & 85.00 & 65.00 & 65.00 & 40.00 & 75.00 & 45.00 & 85.00 & 95.00 \\
\midrule
\multirow{3}{*}{\textbf{GPT-3.5}}
& CS & 70.00 & 00.00 & 00.00 & 20.00 & 10.00 & 05.00 & 10.00 & 10.00 \\
& EN & 75.00 & 00.00 & 00.00 & 10.00 & 15.00 & 0.00 & 05.00 & 00.00 \\
& KOR & 65.00 & 45.00 & 05.00 & 30.00 & 60.00 & 20.00 & 65.00 & 60.00 \\
\midrule
\multirow{3}{*}{\textbf{Llama3-70B}}
& CS & 20.00 & 00.00 & 00.00 & 10.00 & 15.00 & 10.00 & 10.00 & 00.00 \\
& EN & 30.00 & 05.00 & 00.00 & 10.00 & 20.00 & 05.00 & 10.00 & 00.00 \\
& KOR & 60.00 & 50.00 & 00.00 & 40.00 & 70.00 & 35.00 & 55.00 & 60.00 \\
\midrule
\multirow{3}{*}{\textbf{Llama3-8B}}
& CS & 20.00 & 00.00 & 00.00 & 05.00 & 25.00 & 00.00 & 05.00 & 00.00 \\
& EN & 15.00 & 00.00 & 00.00 & 05.00 & 15.00 & 00.00 & 00.00 & 00.00 \\
& KOR & 25.00 & 30.00 & 05.00 & 05.00 & 50.00 & 05.00 & 10.00 & 20.00 \\
\bottomrule

\end{tabular}
}
    \caption{QA performances on open-end QA.}
    \label{tab:openqa}
\end{table*}

%% file: tables/cot_results.tex
\begin{table*}[t]
\small
\centering
\resizebox{1\textwidth}{!}
{
\begin{tabular}{lc cccccccccc}
\toprule 
  Model & &
  \textbf{Economy} & \textbf{General} & \textbf{Geography} & \textbf{History} & \textbf{Law} & \textbf{Politics} & \textbf{Popular} & \textbf{Society} & \textbf{Tradition} & 
  Average \\
\midrule

\multirow{2}{*}{\textbf{GPT-4o}} & 
\cellgray CS & \cellgray \textbf{93.22} & \cellgray \textbf{\green{80.11}} & \cellgray \textbf{69.75} & \cellgray \textbf{\green{76.50}} & \cellgray 49.66 & \cellgray \textbf{\green{92.86}} & \cellgray \textbf{\green{97.56}} & \cellgray \textbf{65.51} & \cellgray \textbf{81.98}  & \cellgray \textbf{78.57}
\\
& EN & 79.66 & 76.14 & 60.14 & 64.96 & \textbf{51.49} & 85.71 & 92.68 & 58.42 & 73.87  & 71.45
\\
\midrule

\multirow{2}{*}{\textbf{GPT-3.5}} &
\cellgray CS & \cellgray \textbf{74.58} & \cellgray 37.50 & \cellgray 39.15 & \cellgray 30.13 & \cellgray 32.41 & \cellgray \textbf{82.14} & \cellgray \textbf{75.61} & \cellgray \textbf{50.50} & \cellgray \textbf{63.06}  & \cellgray \textbf{53.90}
\\
& EN & 69.49 & \textbf{49.43} & \textbf{43.06} & \textbf{34.62} & \textbf{34.02} & 73.81 & 65.85 & 47.03 & 55.41  & 52.52
\\
\midrule

\multirow{2}{*}{\textbf{Claude 3.5}} & 
\cellgray CS & \cellgray \textbf{\green{96.61}} & \cellgray \textbf{78.41} & \cellgray \textbf{\green{78.29}} & \cellgray \textbf{\green{76.50}} & \cellgray \textbf{\green{57.24}} & \cellgray 84.52 & \cellgray \textbf{92.68} & \cellgray \textbf{\green{70.46}} & \cellgray \textbf{\green{86.04}}  & \cellgray \textbf{\green{80.08}}
\\
& EN & 89.83 & 77.84 & 72.60 & 67.52 & 53.79 & \textbf{89.29} & \textbf{92.68} & 62.38 & 81.53  & 76.38
\\
\midrule

\multirow{2}{*}{\textbf{Solar}} & 
\cellgray CS & \cellgray 83.05 & \cellgray \textbf{53.98} & \cellgray \textbf{52.31} & \cellgray \textbf{62.61} & \cellgray \textbf{40.46} & \cellgray \textbf{85.71} & \cellgray \textbf{78.05} & \cellgray \textbf{55.78} & \cellgray \textbf{72.97}  & \cellgray \textbf{64.99}
\\
& EN & \textbf{88.14} & \textbf{53.98} & 47.69 & 37.61 & 39.08 & 76.19 & 70.73 & 51.65 & 65.32  & 58.93
\\
\midrule

\multirow{2}{*}{\textbf{Llama3 70B}} & 
\cellgray CS & \cellgray 76.27 & \cellgray 60.80 & \cellgray 54.45 & \cellgray \textbf{48.29} & \cellgray \textbf{40.46} & \cellgray \textbf{86.90} & \cellgray \textbf{82.93} & \cellgray 55.94 & \cellgray \textbf{71.17}  & \cellgray \textbf{64.13}
\\
& EN & \textbf{79.66} & \textbf{61.36} & \textbf{55.52} & 47.65 & 39.77 & 80.95 & 75.61 & \textbf{56.11} & 68.47  & 62.79
\\
\midrule

\multirow{2}{*}{\textbf{Llama3 8B}} &
\cellgray CS & \cellgray \textbf{76.27} & \cellgray \textbf{39.20} & \cellgray \textbf{40.57} & \cellgray \textbf{38.46} & \cellgray \textbf{33.33} & \cellgray \textbf{72.62} & \cellgray 73.17 & \cellgray 47.03 & \cellgray 56.31  & \cellgray 53.00\\
& EN & 72.88 & 37.50 & \textbf{40.57} & 32.91 & 31.72 & \textbf{72.62} & \textbf{75.61} & \textbf{50.33} & \textbf{59.01}  & 52.57\\
\midrule

\multirow{2}{*}{\textbf{Gemma2 27B}} & 
\cellgray CS & \cellgray 77.97 & \cellgray 51.70 & \cellgray 48.04 & \cellgray \textbf{41.24} & \cellgray 36.78 & \cellgray \textbf{78.57} & \cellgray \textbf{78.05} & \cellgray \textbf{53.63} & \cellgray \textbf{65.32}  & \cellgray \textbf{59.03}
\\
& EN & \textbf{79.66} & \textbf{55.68} & \textbf{50.18} & 36.11 & \textbf{40.46} & 69.05 & 75.61 & 52.15 & 61.26  & 57.80
\\
\midrule

\multirow{2}{*}{\textbf{Gemma2 9B}} & 
\cellgray CS & \cellgray \textbf{76.27} & \cellgray \textbf{50.57} & \cellgray 44.84 & \cellgray \textbf{40.60} & \cellgray 35.63 & \cellgray \textbf{77.38} & \cellgray \textbf{73.17} & \cellgray \textbf{53.14} & \cellgray \textbf{61.71}  & \cellgray \textbf{57.03}
\\
& EN & 67.80 & 44.32 & \textbf{45.20} & 35.68 & \textbf{39.08} & 70.24 & 65.85 & 49.50 & \textbf{61.71}  & 53.26
\\
\bottomrule

\end{tabular}
}

\caption{Knowledge leveraging performances of multilingual LLMs on CS and English settings. \textbf{Bold} indicates higher score between CS and English on each model. \green{\textbf{Green}} indicates the highest score from each domain.
}
\label{tab:cot_results}

\end{table*}

%% file: tables/none.tex
\begin{table*}[h]
\small
\centering
\resizebox{1\textwidth}{!}
{
    \begin{tabular}{lc cccccccccc}
    \toprule
        Model &  & \textbf{Economy} & \textbf{General} & \textbf{Geography} & \textbf{History} & \textbf{Law} & \textbf{Politics} & \textbf{Popular} & \textbf{Society} & \textbf{Tradition} & \textbf{Total} \\ \midrule
        \multirow{2}{*}{\textbf{GPT-4o}} & CS & 0/0 & 0/0 & 0/0 & 0/0 & 0/2 & 0/0 & 0/0 & 0/5 & 0/1 & 0/8 \\ 
        & EN & 0/8 & 0/1 & 1/11 & 0/15 & 2/15 & 1/3 & 0/0 & 1/40 & 0/6 & 5/99 \\ \midrule
        \multirow{2}{*}{\textbf{GPT-3.5}} & CS & 0/0 & \textbf{1/1} & {1/1} & \textbf{2/2} & 0/0 & 0/0 & 0/0 & \textbf{1/1} & 0/0 & \textbf{5/5} \\ 
        & EN & 0/0 & \textbf{3/3} & \textbf{1/1} & \textbf{2/2} & \textbf{1/1} & 0/0 & 0/0 & \textbf{6/6} & 0/0 & \textbf{13/13} \\ \midrule
        \multirow{2}{*}{\textbf{Claude 3.5 Sonnet}} & CS & 0/0 & 0/0 & 0/0 & 0/0 & 0/0 & 0/0 & 0/0 & 0/0 & 0/0 & 0/0 \\ 
        & EN & 0/0 & 0/0 & 0/0 & 0/0 & 0/1 & 0/0 & 0/0 & 0/4 & 0/0 & 0/5 \\ \midrule
        \multirow{2}{*}{\textbf{Solar}} & CS & 0/0 & \textbf{5/5} & \textbf{1/1} & \textbf{2/2} & \textbf{11/11} & 0/0 & 0/0 & \textbf{5/5} & \textbf{1/1} & \textbf{20/20} \\ 
        & EN & 0/0 & \textbf{9/9} & \textbf{8/8} & \textbf{11/11} & \textbf{10/10} & 0/0 & \textbf{1/1} & \textbf{4/4} & \textbf{2/2} & \textbf{35/35} \\ \midrule
        \multirow{2}{*}{\textbf{Llama3 70B}} & CS & \textbf{2/2} & \textbf{6/6} & \textbf{6/6} & \textbf{11/11} & \textbf{3/3} & 0/0 & \textbf{1/1} & \textbf{3/3} & \textbf{2/2} & \textbf{34/34} \\ 
        & EN & \textbf{2/2} & \textbf{7/7} & \textbf{10/10} & \textbf{24/24} & \textbf{12/12} & \textbf{2/2} & 0/0 & \textbf{4/4} & \textbf{2/2} & \textbf{63/63} \\ \midrule
        \multirow{2}{*}{\textbf{Llama3 8B}} & CS & \textbf{1/1} & \textbf{5/5} & \textbf{3/3} & \textbf{9/9} & \textbf{2/2} & \textbf{1/1} & \textbf{1/1} & 0/0 & \textbf{1/1} & \textbf{23/23} \\ 
        & EN & 0/0 & \textbf{4/4} & \textbf{6/6} & \textbf{8/8} & \textbf{5/5} & \textbf{2/2} & 0/0 & \textbf{2/2} & \textbf{2/2} & \textbf{29/29} \\ \midrule
        \multirow{2}{*}{\textbf{Gemma2 27B}} & CS & \textbf{4/4} & \textbf{18/18} & \textbf{7/7} & 26/28 & 22/38 & \textbf{2/2} & 2/3 & 6/9 & 10/14 & 97/123 \\ 
        & EN & \textbf{3/3} & \textbf{28/28} & \textbf{5/5} & 37/38 & 7/15 & \textbf{7/7} & \textbf{1/1} & 13/20 & \textbf{5/5} & 106/122 \\ \midrule
        \multirow{2}{*}{\textbf{Gemma2 9B}} & CS & \textbf{3/3} & \textbf{7/7} & \textbf{12/12} & \textbf{25/25} & 18/19 & \textbf{1/1} & \textbf{2/2} & \textbf{7/7} & \textbf{2/2} & 77/78 \\ 
        & EN & \textbf{9/9} & \textbf{30/30} & \textbf{13/13} & \textbf{35/35} & \textbf{11/11} & \textbf{5/5} & 0/0 & 23/28 & \textbf{9/9} & 135/140 \\ \bottomrule
    \end{tabular}
}

\caption{Counts of None errors. Each cell indicates \# of None errors / \# of errors due to knowledge hallucination. \textbf{Bold} indicates that all errors are caused by hallucination.}
\label{tab:none}

\end{table*}

%% file: figure_latex/llm_faith.tex
\begin{figure*}[!t]
    \centering
    \includegraphics[width=0.8\textwidth]{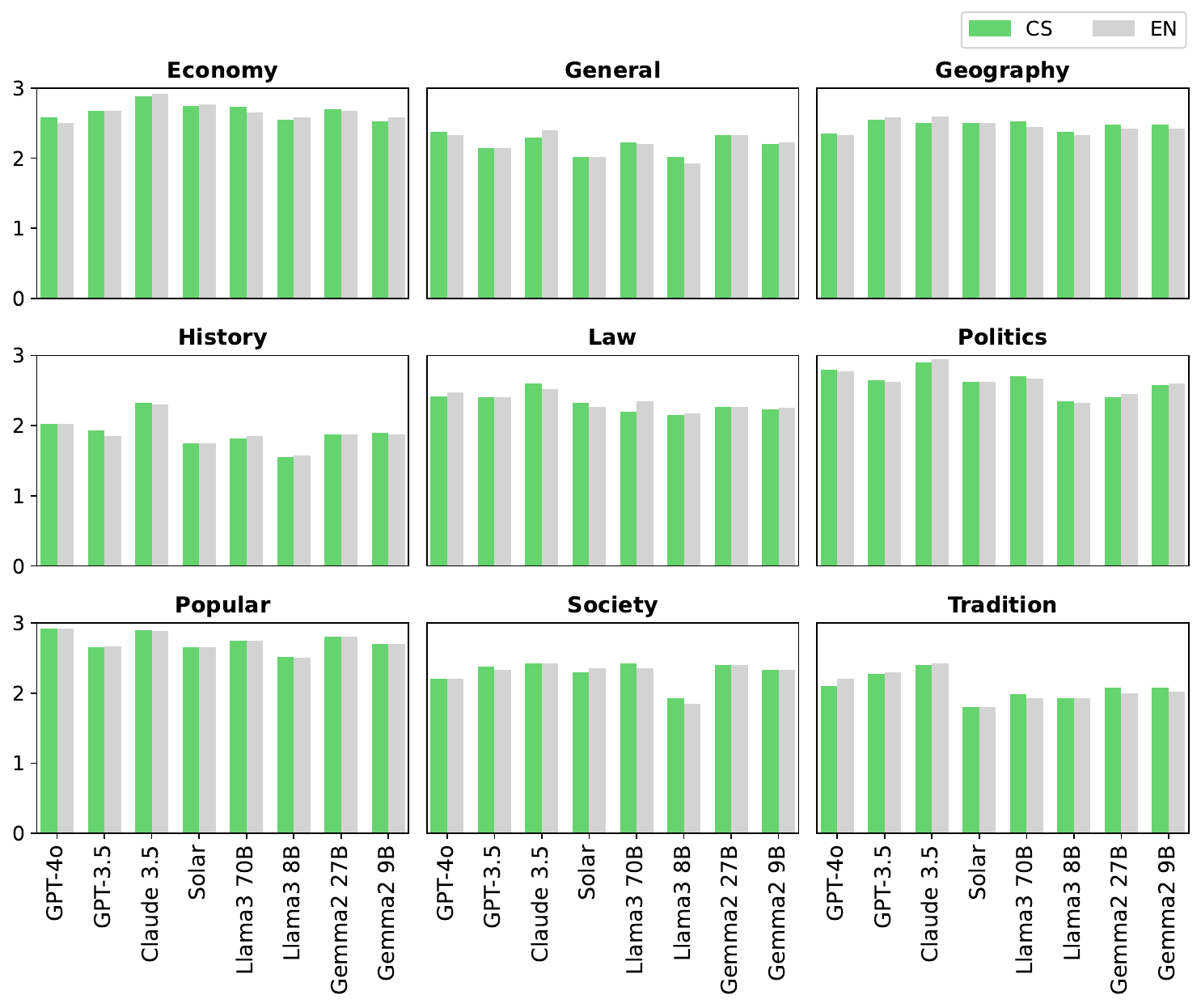}
    
    \caption{LLM-as-a-judge evaluation results on faithfulness between knowledge lists identified from CS and English questions.}
    
    \label{fig:llm_faith} 
    
\end{figure*} 

%% file: figure_latex/llm_help.tex
\begin{figure*}[!t]
    \centering
    \includegraphics[width=0.8\textwidth]{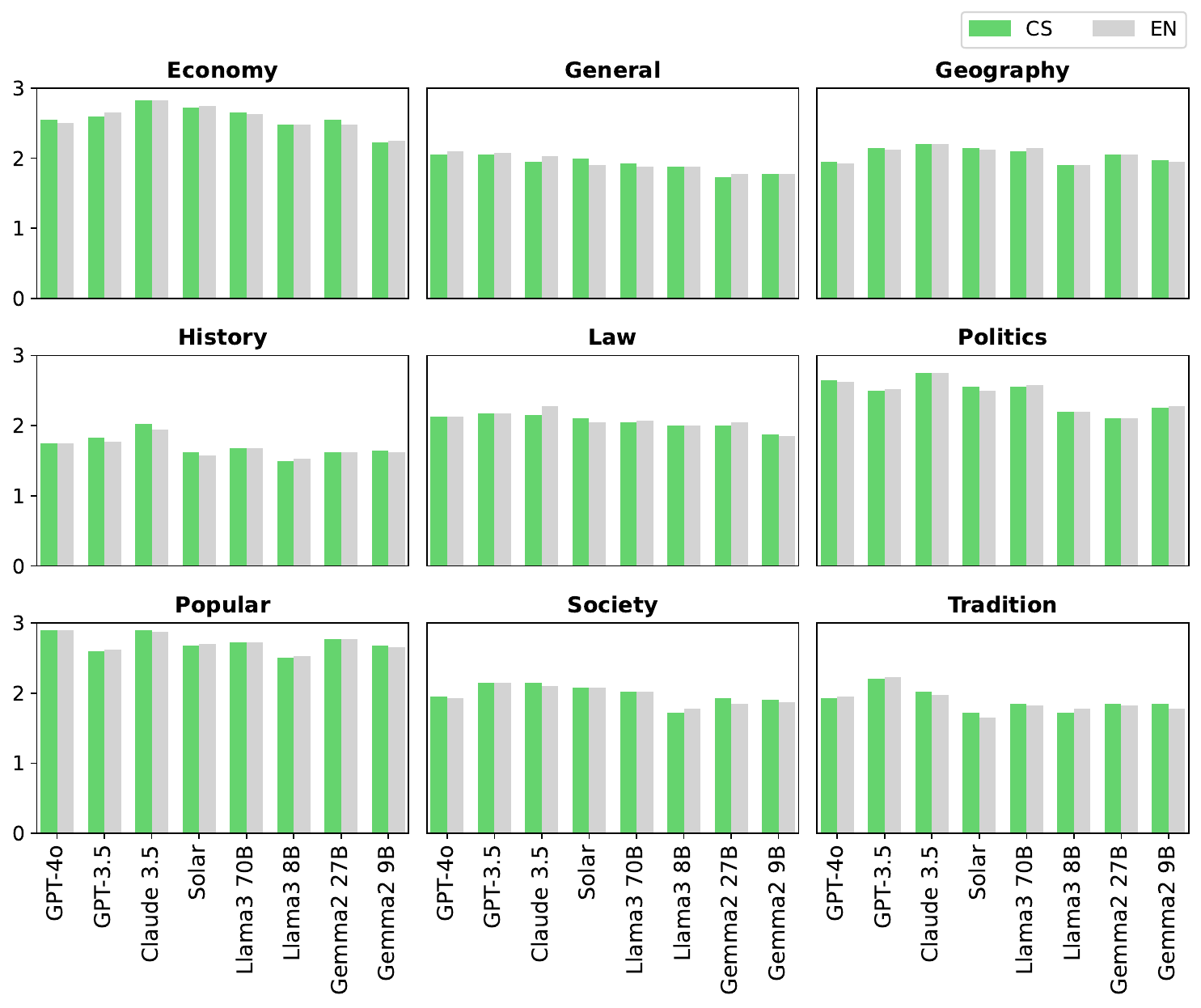}
    
    \caption{LLM-as-a-judge evaluation results on helpfulness between knowledge lists identified from CS and English questions.}
    
    \label{fig:llm_help} 
    
\end{figure*} 

%% file: figure_latex/llm_pair.tex
\begin{figure*}[!t]
    \centering
    \includegraphics[width=0.8\textwidth]{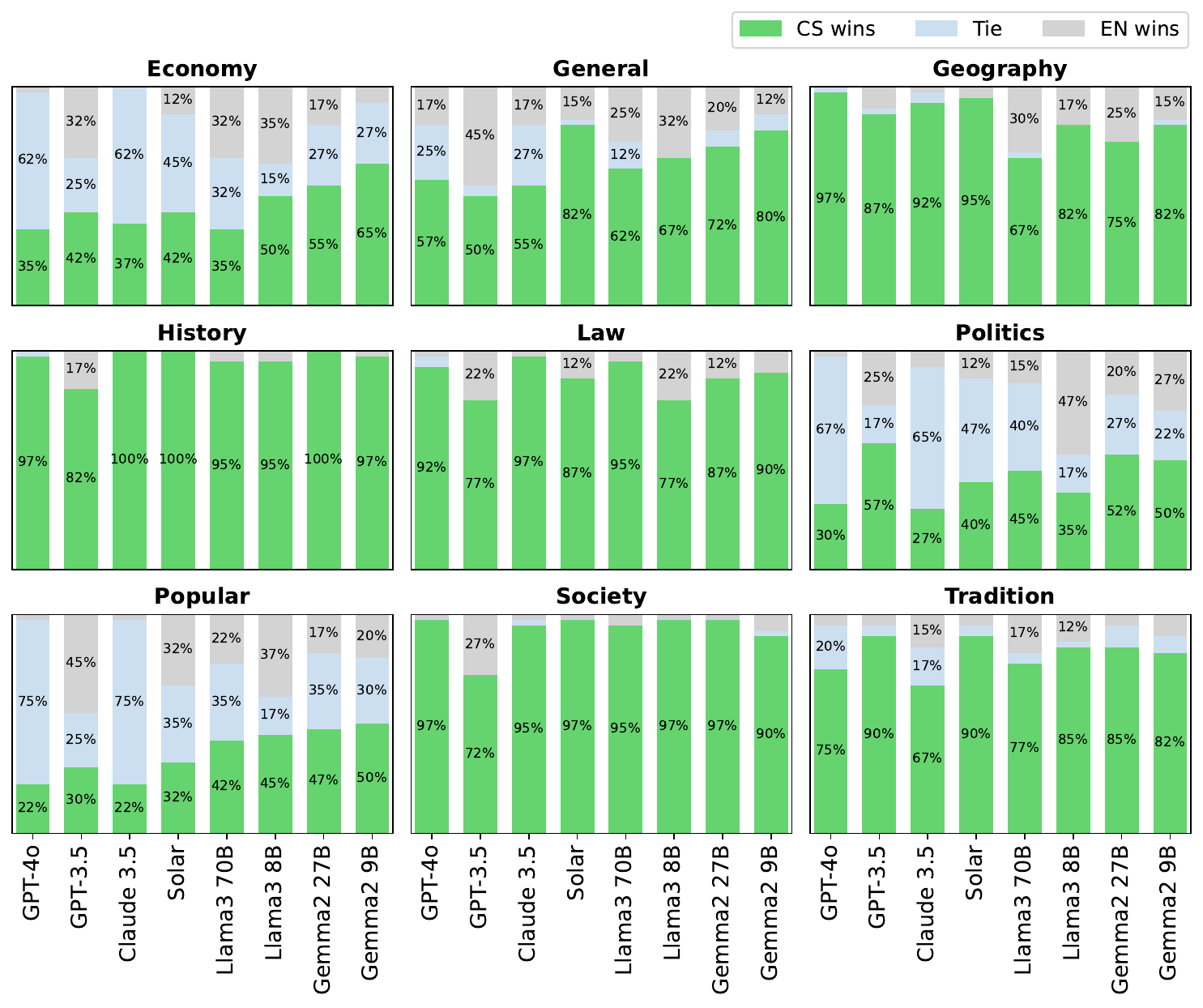}
    
    \caption{LLM-as-a-judge evaluation results on pairwise comparison between knowledge lists identified from CS and English questions.}
    
    \label{fig:llm_pair} 
    
\end{figure*}

%% file: tables/og_kor_result.tex

\begin{table*}[!ht]
\small
\centering
\resizebox{1\textwidth}{!}
{
\begin{tabular}{lc cccccccccc}
\toprule 
  Model & &
  \textbf{Economy} & \textbf{General} & \textbf{Geography} & \textbf{History} & \textbf{Law} & \textbf{Politics} & \textbf{Popular} & \textbf{Society} & \textbf{Tradition} & 
  \textbf{Total} \\
\midrule

\multirow{2}{*}{\textbf{GPT-4o}} & 
 CS &  91.53 &  \textbf{{78.41}} &  69.04 &  74.79 &  55.86 &  \textbf{{90.48}} &  95.12 &  63.70 &  85.14 &  78.23 \\
& KO$_{og}$ & \textbf{94.92} & 76.70 & \textbf{75.09} & \textbf{76.50} & \textbf{58.62} & 89.29 & \textbf{97.56} & \textbf{67.00} & \textbf{85.59} & \textbf{80.14} \\
\midrule

\multirow{2}{*}{\textbf{GPT-3.5}} & 
 CS &  \underline{71.19} &  \textbf{47.73} &  \textbf{44.48} &  \textbf{32.91} &  \textbf{35.40} &  \underline{70.24} &  \textbf{80.49} &  \textbf{49.17} &  \textbf{57.21} &  \textbf{54.31} \\
& KO$_{og}$ & \underline{71.19} & 45.45 & 37.37 & 29.49 & 30.11 & \underline{70.24} & 60.98 & 47.36 & 54.05 & 49.58 \\
\midrule

\multirow{2}{*}{\textbf{Claude 3.5 Sonnet}} & 
 CS &  \textbf{{93.22}} &  \textbf{72.16} &  72.95 &  73.08 &  62.53 &  86.90 &  \underline{{95.12}} &  67.66 &  84.23 &  78.65
\\
& KO$_{og}$ & 89.83 & 71.59 & \textbf{80.78} & \textbf{76.28} & \textbf{66.67} & \textbf{89.29} & \underline{95.12} & \textbf{71.45} & \textbf{87.39} & \textbf{80.93} \\
\midrule

\multirow{2}{*}{\textbf{Solar}} & 
 CS & 83.05 &  \textbf{55.11} &  54.09 &  63.46 &  42.76 &  80.95 &  \textbf{85.37} &  54.29 &  75.23 &  66.03
\\
& KO$_{og}$ & \textbf{84.75} & 51.70 & \textbf{55.87} & \textbf{64.74} & \textbf{43.22} & \textbf{82.14} & 82.93 & \textbf{55.28} & \textbf{76.58} & \textbf{66.36} \\
\midrule

\multirow{2}{*}{\textbf{Llama3 70B}} &
 CS &  79.66 &  51.70 &  50.53 &  49.36 &  \textbf{44.14} &  \textbf{80.95} &  75.61 &  57.43 &  65.77 &  61.68
\\
& KO$_{og}$ & \textbf{86.44} & \textbf{52.27} & \textbf{53.38} & \textbf{51.50} & 41.84 & 77.38 & \textbf{82.93} & \textbf{59.41} & \textbf{68.92} & \textbf{63.79} \\
\midrule

\multirow{2}{*}{\textbf{Llama3 8B}} & 
 CS & 69.49 &  \textbf{40.34} &  36.30 &  35.68 &  \textbf{35.63} &  \textbf{75.00} &  \textbf{73.17} &  45.05 &  54.05 &  51.63
\\
& KO$_{og}$ & \textbf{72.88} & 38.07 & \textbf{37.37} & \textbf{36.75} & 35.40 & 72.62 & 70.73 & \textbf{51.32} & \textbf{55.86} & \textbf{52.33} \\
\midrule

\multirow{2}{*}{\textbf{Gemma2 27B}} & 
 CS &  79.66 &  \textbf{46.02} &  \underline{48.75} &  41.03 &  \textbf{45.29} &  \underline{77.38} &  78.05 &  54.79 &  65.32 &  59.59
\\
& KO$_{og}$ & \textbf{83.05} & 45.45 & \underline{48.75} & \textbf{45.94} & 41.38 & \underline{77.38} & \textbf{80.49} & \textbf{56.60} & \textbf{67.12} & \textbf{60.68} \\
\midrule

\multirow{2}{*}{\textbf{Gemma2 9B}} & 
 CS &  \textbf{79.66} &  \textbf{42.05} &  44.13 &  \textbf{40.17} &  \textbf{41.15} & 73.81 &  \underline{80.49} & 53.30 &  65.77 & 57.84
\\
& KO$_{og}$ & 77.97 & 41.48 & \textbf{45.91} & 38.89 & 40.92 & \textbf{80.95} & \underline{80.49} & \textbf{53.80} & \textbf{66.22} & \textbf{58.51} \\
\bottomrule

\end{tabular}
}

\caption{QA performances of LLMs on CS and original Korean dataset. CS scores are identical with Table~\ref{tab:plain_result}. \textbf{Bold} indicates higher score, and \underline{underline} indicates tie.}
\label{tab:og_kor}
\end{table*}

%% file: tables/case_study1.tex
\begin{table*}[h]
    \small
    \centering
    \resizebox{1\textwidth}{!}{
    \begin{tabular}{p{14cm}}
\toprule
\textbf{Example of Solar from the Tradition domain} \\
\midrule
\textbf{[QUESTION]}\\
The following is a question about 한국의 문화 and 전통.\\
Foreigners want to experience 한국의 전통 풍습 by celebrating the 정월 대보름. Which of the following is not part of the 체험내용?\\ 
(The following is a question about Korean culture and traditions. \\
Foreigners want to experience Korean traditional customs by celebrating the first full moon of the lunar year. Which of the following is not part of the experience?)\\
\\
(1) Eat 오곡밥 made from five types of grains. (Eat five-grain rice made from five types of grains.)\\
(2) Drink 귀밝이술, wishing for better hearing. (Drink ear-brightening potion, wishing for better hearing.)\\
(3) Eat food made with 쑥 and 익모초 to overcome the heat. (Eat food made with mugwort and dried hydrangea to overcome the heat.)\\
(4) Munch on 견과류 like 땅콩, 호두, and 밤, hoping for no skin disease. (Munch on nuts like peanuts, walnuts, and chestnuts, hoping for no skin disease.)\\

\\
\textcolor{green}{\textbf{[Knowledge from CS]}} \\
1. 정월 대보름 (Jeongwol Daeboreum): It is a traditional Korean holiday that takes place on the first full moon of the lunar new year. It is a time for families to gather, participate in various rituals, and enjoy traditional foods.\\
2. 오곡밥 (Ogokbap): A traditional rice cake made with five grains (rice, barley, soybeans, sesame seeds, and black beans) that is eaten during 정월 대보름 for good health and longevity.\\
3. 귀밝이술 (Gwi-bal-i-sul): A traditional Korean drink made from makgeolli (rice wine) that is consumed on 정월 대보름 to improve hearing and communication skills.\\
4. 쑥과 익모초 (Perilla and Mugwort): These herbs are used in traditional Korean medicine and are believed to have healing properties. While they may be used in some traditional Korean dishes, they are not specifically associated with 정월 대보름.\\
5. 견과류 (Nuts): Eating nuts like 땅콩 (peanuts), 호두 (walnuts), and 밤 (chestnuts) is a common practice during 정월 대보름 as they are believed to bring good luck and prevent skin diseases.\\
\\
\textcolor{blue}{[ANSWER]}\\
(3)
\\\\
\textbf{\textcolor{gray}{\textbf{[Knowledge from EN]}}} \\
1. The first full moon of the lunar year is celebrated in Korea as a traditional holiday called "Dano."\\
2. One of the customs during Dano is eating "ongchimsik," a rice dish made with five grains.\\
3. Another custom is drinking "suriseong," a herbal drink believed to improve hearing.\\
4. Eating "yeot," a dish made with mugwort and dried hydrangea, is also a traditional practice during Dano to help overcome the heat.\\
5. Nuts like peanuts, walnuts, and chestnuts are often consumed during Dano as a symbol of good health and fortune.\\
6. Skin disease is not directly related to the traditional customs of Dano.\\
\\
\textcolor{red}{[ANSWER]}\\
(4)
    ... \\

\bottomrule
    \end{tabular}
    }
    \caption{An example of knowledge activation by CS and English questions from the Tradition domain on Solar. The text in parentheses shows English translation.}
    \label{tab:case_study1}
\end{table*}

%% file: tables/case_study2.tex
\begin{table*}[h]
    \small
    \centering
    \resizebox{1\textwidth}{!}{
    \begin{tabular}{p{14cm}}
\toprule
\textbf{Example of Gemma2 9B from the Law domain} \\
\midrule
\textbf{[QUESTION]}\\ 
What is the correct legal 판단 on the following case?\\
17세인 A, B, and C는 entered into a 계약 with 판매업자 D to purchase an expensive 노트북. A did not obtain 부모의 동의, while both B and C presented 부모의 동의서, but C forged the consent form. At the time of the contract, D did not know that A was a 미성년자. *Parents of A, B, and C are all 법정 대리인.\\
(What is the correct legal judgment on the following case?\\
17-year-olds A, B, and C entered into a contract with seller D to purchase an expensive laptop. A did not obtain parental consent, while both B and C presented parental consent forms, but C forged the consent form. At the time of the contract, D did not know that A was a minor. * Parents of A, B, and C are all legal representatives.)\\
\\
(1) B's 부모 can cancel the 계약. (B's parents can cancel the contract.)\\
(2) The 부모 of A and C can both cancel the 계약. (The parents of A and C can both cancel the contract.)\\
(3) B cannot cancel the 계약 unlike A. (B cannot cancel the contract unlike A.)\\
(4) D cannot withdraw the 의사 표시 to conclude a 계약 with A. (D cannot withdraw the expression of intention to conclude a contract with A.)\\
(5) D can demand a prompt response from A regarding the 취소 of the 계약. (D can demand a prompt response from A regarding the cancellation of the contract.)\\
\\
\textcolor{green}{\textbf{[Knowledge from CS]}} \\
1. **미성년자 계약의 효력**: 대한민국 민법상 14세 미만의 자는 계약력이 없으며, 14세 이상 19세 미만의 자는 부모의 동의가 없는 계약은 효력이 없다.\\
2. **부모의 동의**: 미성년자의 계약은 부모의 동의가 있어야 효력이 있다. 동의서가 위조된 경우, 계약은 무효이다.\\
3. **계약의 취소**: 계약의 당사자는 계약을 취소할 수 있다. \\
4. **의사표시의 취소**: 의사표시는 당사자의 의사에 따라 취소될 수 있다. \\
\\
\textcolor{red}{[ANSWER]} (Correct answer is (3).)\\
(2)
\\\\
\textbf{\textcolor{gray}{\textbf{[Knowledge from EN]}}} \\
1.  In most jurisdictions, minors (typically under 18 years old) lack the legal capacity to enter into binding contracts. \\
2.  Contracts entered into by minors are generally voidable at the option of the minor.\\
3.  Parental consent can sometimes affect the validity of a contract entered into by a minor, depending on the specific laws of the jurisdiction.\\
4.  Fraudulent misrepresentation, such as forging a parental consent form, can invalidate a contract. \\
\\
\textcolor{red}{[ANSWER]}\\
(2)\\ 

\bottomrule
    \end{tabular}
    }
    \caption{An example of knowledge activation by CS and English questions from the Law domain on Gemma2 9B. The text in parentheses shows English translation.}
    \label{tab:case_study2}
\end{table*}

%% file: tables/prompt_dataset.tex
\begin{table*}[h]
    \small
    \centering
    \begin{tabular}{p{14cm}}
    \toprule
    \textbf{Prompt for generating CS candidates} \\
    \midrule
You are a bilingual who can speak both English and Korean fluently. I will give you a Korean and English (<Korean>, <English>) pair. They are semantically the same. Your job is to write Korean-English code-switching text with certain switching level by mixing Korean and English text. Example 1 is an example of code-switched text in different levels of code-switching. Use Example 1 as reference to understand the level of code-switching. Read the instructions carefully and solve the Task.\\
\\
Instructions:\\
- Maintain English word order, that is, Subject-Verb-Object.\\
- Find semantically important given nouns and noun phrases from the text, and change \texttt{\{level\}} percent of them to Korean. \\
- Keep functional words in English.\\
- Keep the indicators such as (가), (나), ㄱ, ㄴ, 갑, 을 in Korean.\\
\\
{[Example 1]}\\
<Korean> \\
제주도는 점성이 작고 유동성이 큰 마그마가 여러 차례 분출하여 형성된 방패 모양의 화산섬이다. 하지만 한라산의 정상부는 종 모양의 화산으로 이루어져 있으며, 산허리에는 오름으로 불리는 기생화산이 많이 형성되어 있다.\\
\\
<English> \\
Jeju Island is a shield-shaped volcanic island formed by multiple eruptions of small-sized and highly fluid magma. However, the top of Hallasan Mountain consists of a cone-shaped volcano, and many parasitic volcanoes called Oreum are formed on the hillsides.\\
\\
<Code-switch with 30 percent of Korean>\\
Jeju Island is a shield-shaped 화산섬 formed by multiple eruptions of small-sized and highly fluid magma. However, the top of 한라산 consists of a cone-shaped volcano, and many 기생화산 called 오름 are formed on the hillsides.\\
\\
<Code-switch with 50 percent of Korean>\\
Jeju Island is a 방패 모양의 화산섬 formed by multiple eruptions of small-sized and highly fluid 마그마. However, the top of 한라산 consists of a cone-shaped 화산, and many 기생화산 called 오름 are formed on the hillsides.\\
\\
<Code-switch with 70 percent of Korean>\\
제주도 is a 방패 모양의 화산섬 formed by multiple eruptions of 크기가 작고 유동성이 큰 마그마. However, the top of 한라산 Mountain consists of a 종 모양의 화산, and many 기생화산 called 오름 are formed on the 산허리.
\\
<Code-switch with 90 percent of Korean>\\
제주도 is a shield-shaped 화산섬 formed by multiple 분출 of small-sized and 유동성이 큰 마그마. However, the 정상부 of 한라산 consists of a cone-shaped 화산, and many 기생화산 called 오름 are formed on the 산허리.
\\
\\
{[Task]}\\
<Korean>\\
\texttt{\{question\}}\\
\\
<English>\\
\texttt{\{translation\}}\\
\\
<Code-Switch>\\
 \\\bottomrule
    \end{tabular}
    \caption{Prompt for generating code-switched text candidates in diffferent levels.}
    \label{tab:prompt_dataset}
\end{table*}

%% file: tables/prompt_plain.tex
\begin{table*}[h]
    \small
    \centering
    \begin{tabular}{p{14cm}}
    \toprule
    \textbf{Prompt for QA (CS)} \\
    \midrule
You will be given a question and choices about Korea. The text are written in English-Korean code-switching, where matrix language is English and semantically important Korean words are embedded into English sentence. Your job is to answer the question. Read the [QUESTION] and choose the most appropriate answer from [CHOICES]. Only write your answer number in parentheses, like (1). Do not repeat the question or choice. \\
Use Example 1 as a reference to answer Example 2.\\
\\
<Example 1> \\
{[QUESTION]} \\
Which city is the 수도 of 한국? \\
\\
{[CHOICES]}\\
(1) 뉴욕 (New York)\\
(2) 서울 (Seoul)\\
(3) 파리 (Paris) \\
(4) 도쿄 (Tokyo)\\
\\
{[ANSWER]}\\
(2)\\
\\
<Example 2>\\
{[QUESTION]}\\
\{question\}\\
{[ANSWER]}\\

\bottomrule
    \end{tabular}
    \caption{Prompt for QA inference.}
    \label{tab:prompt_plain}
\end{table*}

%% file: tables/prompt_identify.tex
\begin{table*}[h]
    \small
    \centering
    \begin{tabular}{p{14cm}}
    \toprule
    \textbf{Prompt for Knowledge Identification} \\
    \midrule
You are a bilingual who is fluent in both Korean and English, and is knowledgeable about South Korea. You will be given a multiple choice question about South Korea. The text are written in English-Korean code-switching, where matrix language is English and semantically important Korean words are embedded into English sentence. Your job is to follow the instructions and write a list of knowledge that is necessary to know for solving the question correctly. \\
\\
Instructions:\\
- Write a list of factual knowledge that are required for solving the question. Try to write each knowledge in one or two sentences. You can write in whichever language you can explain better, either Korean or English. Start this task with [KNOWLEDGE] tag.\\
- Only write knowledge that you definitely know. Do not write incorrect information. \\
- Do not repeat input text in your response. Do not generate new question. Stick to input text that is given to you.\\
\\

I will give you an example for reference. \\
<<Example 1>>\\
{[QUESTION]}\\
Read the following question and choose the most appropriate answer. Who is the person who greatly defeated the soldiers of the 당나라 in the 안시성 싸움?\\
\\
{[CHOICES]}\\
(1) 양만춘\\
(2) 서희\\
(3) 김유신\\
(4) 강감찬\\
(5) 윤관\\
\\
{[KNOWLEDGE]}\\
1. 안시성 싸움 (Siege of Ansi): 안시성 싸움 (645 AD) was a famous military conflict between 고구려 and the 당 Dynasty. 고구려, under the leadership of 양만춘 (Yang Man-chun), successfully defended the 안시성 against the powerful 당 forces led by Emperor 태종.  \\
2. 양만춘 (Yang Man-chun): He was the general who commanded the defense of 안시성, playing a key role in defeating the 당나라 army.  \\
3. 서희 (Seo Hee): A 고려 diplomat famous for negotiating with the 거란 to avoid invasion, but not involved in the 안시성 싸움.  \\
4. 김유신 (Kim Yu-shin): A general from the 신라 Kingdom, instrumental in the unification of the 한반도, but not involved in this specific battle.  \\
5. 강감찬 (Gang Gam-chan): A 고려 military commander known for his victory over the 거란 in the 귀주대첩, unrelated to 안시성.  \\
6. 윤관 (Yun Gwan): A 고려 general famous for his campaigns against the Jurchen, unrelated to the 한반도.\\

\\
Now solve this.\\
<<Example 2>>\\
{[QUESTION]}\\
\{question\}\\
\\
{[CHOICES]}\\
\{choices\}\\

 \\ \bottomrule
    \end{tabular}
    \caption{Prompt for Knowledge Identification task.}
    \label{tab:prompt_identify}
\end{table*}

%% file: tables/prompt_leveraging.tex
\begin{table*}[h]
    \small
    \centering
    \begin{tabular}{p{14cm}}
    \toprule
    \textbf{Prompt for Knowledge Leveraging} \\
    \midrule
You are a bilingual who is fluent in both Korean and English, and is knowledgeable about South Korea. You will be given a multiple choice question and a list of knowledge that are relevant to the question. The text are written in English-Korean code-switching, where matrix language is English and semantically important Korean words are embedded into English sentence. Your job is to follow the instructions and select one choice from [CHOICES].\\
\\
Instructions:\\
- Using given [KNOWLEDGE], explain concisely what and why you think is the answer. You can write in whichever language you can explain better, either Korean or English. Start this task with [EXPLANATION] tag.\\
- Choose your final choice from [CHOICES]. The answer is one of the [CHOICES], so do not say 'none of the above'. You must write a index number in parentheses, like (1). Start this task with [ANSWER] tag.\\
- Do not repeat input text in your response. Do not generate new question. Stick to input text that is given to you.\\

\\
I will give you an example for reference. \\
<<Example 1>>\\
{[QUESTION]}\\
Read the following question and choose the most appropriate answer. Who is the person who greatly defeated the soldiers of the 당나라 in the 안시성 싸움?\\
\\
{[CHOICES]}\\
(1) 양만춘\\
(2) 서희\\
(3) 김유신\\
(4) 강감찬\\
(5) 윤관\\
\\
{[KNOWLEDGE]}\\
1. 안시성 싸움 (Siege of Ansi): 안시성 싸움 (645 AD) was a famous military conflict between 고구려 and the 당 Dynasty. 고구려, under the leadership of 양만춘 (Yang Man-chun), successfully defended the 안시성 against the powerful 당 forces led by Emperor 태종.  \\
2. 양만춘 (Yang Man-chun): He was the general who commanded the defense of 안시성, playing a key role in defeating the 당나라 army.  \\
3. 서희 (Seo Hee): A 고려 diplomat famous for negotiating with the 거란 to avoid invasion, but not involved in the 안시성 싸움.  \\
4. 김유신 (Kim Yu-shin): A general from the 신라 Kingdom, instrumental in the unification of the 한반도, but not involved in this specific battle.  \\
5. 강감찬 (Gang Gam-chan): A 고려 military commander known for his victory over the 거란 in the 귀주대첩, unrelated to 안시성.  \\
6. 윤관 (Yun Gwan): A 고려 general famous for his campaigns against the Jurchen, unrelated to the 한반도.\\
\\
{[EXPLANATION]}\\
The question specifically asks about the 안시성 싸움 (Siege of Ansi) and who defeated the 당나라 soldiers in that battle. Based on historical facts, the leader who played a key role in defending 안시성 and defeating the 당나라 army was 양만춘 (Yang Man-chun).\\
\\
{[ANSWER]}\\
(1)\\
\\
Now solve this.\\
<<Example 2>>\\
{[QUESTION]}\\
\{question\}\\

{[CHOICES]}\\
\{choices\}\\
\\
{[KNOWLEDGE]}\\
\{knowledge\}\\

\\ \bottomrule
    \end{tabular}
    \caption{Prompt for Knowledge Leveraging task.}
    \label{tab:prompt_leveraging}
\end{table*}

%% file: _main.bbl
\begin{thebibliography}{40}
\providecommand{\natexlab}[1]{#1}

\bibitem[{Aguilar et~al.(2020)Aguilar, Kar, and Solorio}]{aguilar-etal-2020-lince}
Gustavo Aguilar, Sudipta Kar, and Thamar Solorio. 2020.
\newblock \href {https://aclanthology.org/2020.lrec-1.223} {{L}in{CE}: A centralized benchmark for linguistic code-switching evaluation}.
\newblock In \emph{Proceedings of the Twelfth Language Resources and Evaluation Conference}, pages 1803--1813, Marseille, France. European Language Resources Association.

\bibitem[{Anthropic(2024)}]{claude3.5}
Anthropic. 2024.
\newblock \href {https://www.anthropic.com/news/claude-3-5-sonnet} {Claude 3.5 sonnet}.

\bibitem[{Artetxe et~al.(2023)Artetxe, Goswami, Bhosale, Fan, and Zettlemoyer}]{artetxe-etal-2023-revisiting}
Mikel Artetxe, Vedanuj Goswami, Shruti Bhosale, Angela Fan, and Luke Zettlemoyer. 2023.
\newblock \href {https://doi.org/10.18653/v1/2023.emnlp-main.399} {Revisiting machine translation for cross-lingual classification}.
\newblock In \emph{Proceedings of the 2023 Conference on Empirical Methods in Natural Language Processing}, pages 6489--6499, Singapore. Association for Computational Linguistics.

\bibitem[{Barei\ss{} et~al.(2024)Barei\ss{}, Klinger, and Barnes}]{baresiss}
Patrick Barei\ss{}, Roman Klinger, and Jeremy Barnes. 2024.
\newblock \href {https://doi.org/10.1145/3589335.3651902} {English prompts are better for nli-based zero-shot emotion classification than target-language prompts}.
\newblock In \emph{Companion Proceedings of the ACM Web Conference 2024}, WWW '24, page 1318–1326, New York, NY, USA. Association for Computing Machinery.

\bibitem[{Chen et~al.(2024)Chen, Ji, Bogoychev, Kutuzov, Haddow, and Heafield}]{chen-etal-2024-monolingual}
Pinzhen Chen, Shaoxiong Ji, Nikolay Bogoychev, Andrey Kutuzov, Barry Haddow, and Kenneth Heafield. 2024.
\newblock \href {https://aclanthology.org/2024.findings-eacl.90/} {Monolingual or multilingual instruction tuning: Which makes a better alpaca}.
\newblock In \emph{Findings of the Association for Computational Linguistics: EACL 2024}, pages 1347--1356, St. Julian{'}s, Malta. Association for Computational Linguistics.

\bibitem[{Do{\u{g}}ru{\"o}z et~al.(2021)Do{\u{g}}ru{\"o}z, Sitaram, Bullock, and Toribio}]{dogruoz-etal-2021-survey}
A.~Seza Do{\u{g}}ru{\"o}z, Sunayana Sitaram, Barbara~E. Bullock, and Almeida~Jacqueline Toribio. 2021.
\newblock \href {https://doi.org/10.18653/v1/2021.acl-long.131} {A survey of code-switching: Linguistic and social perspectives for language technologies}.
\newblock In \emph{Proceedings of the 59th Annual Meeting of the Association for Computational Linguistics and the 11th International Joint Conference on Natural Language Processing (Volume 1: Long Papers)}, pages 1654--1666, Online. Association for Computational Linguistics.

\bibitem[{Dubey et~al.(2024)Dubey, Jauhri, Pandey, Kadian, Al-Dahle, Letman, Mathur, Schelten, Yang, Fan et~al.}]{dubey2024llama3herdmodels}
Abhimanyu Dubey, Abhinav Jauhri, Abhinav Pandey, Abhishek Kadian, Ahmad Al-Dahle, Aiesha Letman, Akhil Mathur, Alan Schelten, Amy Yang, Angela Fan, et~al. 2024.
\newblock \href {https://arxiv.org/abs/2407.21783} {The llama 3 herd of models}.
\newblock \emph{Preprint}, arXiv:2407.21783.

\bibitem[{{Gemma Team}(2024)}]{gemma2}
{Gemma Team}. 2024.
\newblock \href {https://storage.googleapis.com/deepmind-media/gemma/gemma-2-report.pdf} {Gemma 2: Improving open language models at a practical size.}

\bibitem[{Heredia and Altarriba(2001)}]{why-code-switch}
Roberto Heredia and Jeanette Altarriba. 2001.
\newblock \href {https://doi.org/10.1111/1467-8721.00140} {Bilingual language mixing: Why do bilinguals code-switch?}
\newblock \emph{Current Directions in Psychological Science - CURR DIRECTIONS PSYCHOL SCI}, 10:164--168.

\bibitem[{Huzaifah et~al.(2024)Huzaifah, Zheng, Chanpaisit, and Wu}]{huzaifah-etal-2024-evaluating}
Muhammad Huzaifah, Weihua Zheng, Nattapol Chanpaisit, and Kui Wu. 2024.
\newblock \href {https://aclanthology.org/2024.lrec-main.565} {Evaluating code-switching translation with large language models}.
\newblock In \emph{Proceedings of the 2024 Joint International Conference on Computational Linguistics, Language Resources and Evaluation (LREC-COLING 2024)}, pages 6381--6394, Torino, Italia. ELRA and ICCL.

\bibitem[{Joshi et~al.(2020)Joshi, Santy, Budhiraja, Bali, and Choudhury}]{joshi-etal-2020-state}
Pratik Joshi, Sebastin Santy, Amar Budhiraja, Kalika Bali, and Monojit Choudhury. 2020.
\newblock \href {https://doi.org/10.18653/v1/2020.acl-main.560} {The state and fate of linguistic diversity and inclusion in the {NLP} world}.
\newblock In \emph{Proceedings of the 58th Annual Meeting of the Association for Computational Linguistics}, pages 6282--6293, Online. Association for Computational Linguistics.

\bibitem[{Khanuja et~al.(2020)Khanuja, Dandapat, Srinivasan, Sitaram, and Choudhury}]{khanuja-etal-2020-gluecos}
Simran Khanuja, Sandipan Dandapat, Anirudh Srinivasan, Sunayana Sitaram, and Monojit Choudhury. 2020.
\newblock \href {https://doi.org/10.18653/v1/2020.acl-main.329} {{GLUEC}o{S}: An evaluation benchmark for code-switched {NLP}}.
\newblock In \emph{Proceedings of the 58th Annual Meeting of the Association for Computational Linguistics}, pages 3575--3585, Online. Association for Computational Linguistics.

\bibitem[{Kim et~al.(2024{\natexlab{a}})Kim, Park, Kim, Lee, Song, Kim, Kim, Kim, Lee, Kim, Ahn, Yang, Lee, Park, Gim, Cha, Lee, and Kim}]{kim2024solar107bscalinglarge}
Dahyun Kim, Chanjun Park, Sanghoon Kim, Wonsung Lee, Wonho Song, Yunsu Kim, Hyeonwoo Kim, Yungi Kim, Hyeonju Lee, Jihoo Kim, Changbae Ahn, Seonghoon Yang, Sukyung Lee, Hyunbyung Park, Gyoungjin Gim, Mikyoung Cha, Hwalsuk Lee, and Sunghun Kim. 2024{\natexlab{a}}.
\newblock \href {https://arxiv.org/abs/2312.15166} {Solar 10.7b: Scaling large language models with simple yet effective depth up-scaling}.
\newblock \emph{Preprint}, arXiv:2312.15166.

\bibitem[{Kim et~al.(2024{\natexlab{b}})Kim, Suk, Oh, Yoo, Thorne, and Oh}]{kim-etal-2024-click}
Eunsu Kim, Juyoung Suk, Philhoon Oh, Haneul Yoo, James Thorne, and Alice Oh. 2024{\natexlab{b}}.
\newblock \href {https://aclanthology.org/2024.lrec-main.296} {{CLI}c{K}: A benchmark dataset of cultural and linguistic intelligence in {K}orean}.
\newblock In \emph{Proceedings of the 2024 Joint International Conference on Computational Linguistics, Language Resources and Evaluation (LREC-COLING 2024)}, pages 3335--3346, Torino, Italia. ELRA and ICCL.

\bibitem[{Landis and Koch(1977)}]{landis-koch}
J.~Richard Landis and Gary~G. Koch. 1977.
\newblock \href {http://www.jstor.org/stable/2529310} {The measurement of observer agreement for categorical data}.
\newblock \emph{Biometrics}, 33(1):159--174.

\bibitem[{Lee et~al.(2024)Lee, Jung, and Hwang}]{lee-etal-2024-commit}
Jaeseong Lee, YeonJoon Jung, and Seung-won Hwang. 2024.
\newblock \href {https://doi.org/10.18653/v1/2024.findings-naacl.198} {{COMMIT}: Code-mixing {E}nglish-centric large language model for multilingual instruction tuning}.
\newblock In \emph{Findings of the Association for Computational Linguistics: NAACL 2024}, pages 3130--3137, Mexico City, Mexico. Association for Computational Linguistics.

\bibitem[{Li et~al.(2024)Li, Huang, Ching, Dai, and Chen}]{li-etal-2024-prealign}
Jiahuan Li, Shujian Huang, Aarron Ching, Xinyu Dai, and Jiajun Chen. 2024.
\newblock \href {https://doi.org/10.18653/v1/2024.emnlp-main.572} {{P}re{A}lign: Boosting cross-lingual transfer by early establishment of multilingual alignment}.
\newblock In \emph{Proceedings of the 2024 Conference on Empirical Methods in Natural Language Processing}, pages 10246--10257, Miami, Florida, USA. Association for Computational Linguistics.

\bibitem[{Li et~al.(2023)Li, Zhu, Lu, and Yin}]{li-etal-2023-synthetic}
Zhuoyan Li, Hangxiao Zhu, Zhuoran Lu, and Ming Yin. 2023.
\newblock \href {https://doi.org/10.18653/v1/2023.emnlp-main.647} {Synthetic data generation with large language models for text classification: Potential and limitations}.
\newblock In \emph{Proceedings of the 2023 Conference on Empirical Methods in Natural Language Processing}, pages 10443--10461, Singapore. Association for Computational Linguistics.

\bibitem[{Maia~Polo et~al.(2024)Maia~Polo, Weber, Choshen, Sun, Xu, and Yurochkin}]{pmlr-v235-maia-polo24a}
Felipe Maia~Polo, Lucas Weber, Leshem Choshen, Yuekai Sun, Gongjun Xu, and Mikhail Yurochkin. 2024.
\newblock \href {https://proceedings.mlr.press/v235/maia-polo24a.html} {tiny{B}enchmarks: evaluating {LLM}s with fewer examples}.
\newblock In \emph{Proceedings of the 41st International Conference on Machine Learning}, volume 235 of \emph{Proceedings of Machine Learning Research}, pages 34303--34326. PMLR.

\bibitem[{Myers-Scotton(1997)}]{myers1997duelling}
C.~Myers-Scotton. 1997.
\newblock \href {https://books.google.co.kr/books?id=NuYdnTyKkdQC} {\emph{Duelling Languages: Grammatical Structure in Codeswitching}}.
\newblock Clarendon Press.

\bibitem[{Nigatu et~al.(2024)Nigatu, Tonja, Rosman, Solorio, and Choudhury}]{nigatu-etal-2024-zenos}
Hellina~Hailu Nigatu, Atnafu~Lambebo Tonja, Benjamin Rosman, Thamar Solorio, and Monojit Choudhury. 2024.
\newblock \href {https://doi.org/10.18653/v1/2024.emnlp-main.983} {The zeno{'}s paradox of `low-resource' languages}.
\newblock In \emph{Proceedings of the 2024 Conference on Empirical Methods in Natural Language Processing}, pages 17753--17774, Miami, Florida, USA. Association for Computational Linguistics.

\bibitem[{OpenAI(2023)}]{openai2024gpt4technicalreport}
OpenAI. 2023.
\newblock \href {https://arxiv.org/abs/2303.08774} {Gpt-4 technical report}.
\newblock \emph{Preprint}, arXiv:2303.08774.

\bibitem[{Pacchiardi et~al.(2024)Pacchiardi, Cheke, and Hernández-Orallo}]{pacchiardi2024100instancesneedpredicting}
Lorenzo Pacchiardi, Lucy~G. Cheke, and José Hernández-Orallo. 2024.
\newblock \href {https://arxiv.org/abs/2409.03563} {100 instances is all you need: predicting the success of a new llm on unseen data by testing on a few instances}.
\newblock \emph{Preprint}, arXiv:2409.03563.

\bibitem[{Park et~al.(2024)Park, Kim, Kim, Cho, Kim, Lee, Kim, and Lee}]{open-ko-llm}
Chanjun Park, Hyeonwoo Kim, Dahyun Kim, Seonghwan Cho, Sanghoon Kim, Sukyung Lee, Yungi Kim, and Hwalsuk Lee. 2024.
\newblock Open ko-llm leaderboard: Evaluating large language models in korean with ko-h5 benchmark.
\newblock In \emph{The 62nd Annual Meeting of the Association for Computational Linguistics (ACL 2024)}.

\bibitem[{Park and Yun(2021)}]{kor-eng-grammar}
Eunsun Park and Hongoak Yun. 2021.
\newblock The grammatical constraint and grammatical encoding of {Korean-English} code switching.
\newblock \emph{The Journal of Mirae English Language and Literature}, 26(1):177--204.

\bibitem[{Poplack(1980)}]{poplack1980}
Shana Poplack. 1980.
\newblock \href {https://doi.org/10.1515/ling.1980.18.7-8.581} {Sometimes i’ll start a sentence in spanish y termino en espaÑol: toward a typology of code-switching 1}.
\newblock \emph{Linguistics}, 18:581--618.

\bibitem[{Pratapa et~al.(2018)Pratapa, Bhat, Choudhury, Sitaram, Dandapat, and Bali}]{pratapa-etal-2018-language}
Adithya Pratapa, Gayatri Bhat, Monojit Choudhury, Sunayana Sitaram, Sandipan Dandapat, and Kalika Bali. 2018.
\newblock \href {https://doi.org/10.18653/v1/P18-1143} {Language modeling for code-mixing: The role of linguistic theory based synthetic data}.
\newblock In \emph{Proceedings of the 56th Annual Meeting of the Association for Computational Linguistics (Volume 1: Long Papers)}, pages 1543--1553, Melbourne, Australia. Association for Computational Linguistics.

\bibitem[{Rizvi et~al.(2021)Rizvi, Srinivasan, Ganu, Choudhury, and Sitaram}]{rizvi-etal-2021-gcm}
Mohd Sanad~Zaki Rizvi, Anirudh Srinivasan, Tanuja Ganu, Monojit Choudhury, and Sunayana Sitaram. 2021.
\newblock \href {https://doi.org/10.18653/v1/2021.eacl-demos.24} {{GCM}: A toolkit for generating synthetic code-mixed text}.
\newblock In \emph{Proceedings of the 16th Conference of the European Chapter of the Association for Computational Linguistics: System Demonstrations}, pages 205--211, Online. Association for Computational Linguistics.

\bibitem[{Salaam et~al.(2022)Salaam, Dernoncourt, Bui, Rawat, and Yoon}]{salaam-etal-2022-offensive}
Cesa Salaam, Franck Dernoncourt, Trung Bui, Danda Rawat, and Seunghyun Yoon. 2022.
\newblock \href {https://aclanthology.org/2022.coling-1.575} {Offensive content detection via synthetic code-switched text}.
\newblock In \emph{Proceedings of the 29th International Conference on Computational Linguistics}, pages 6617--6624, Gyeongju, Republic of Korea. International Committee on Computational Linguistics.

\bibitem[{Shankar et~al.(2024)Shankar, Jyothi, and Bhattacharyya}]{shankar-etal-2024-icm}
Bhavani Shankar, Preethi Jyothi, and Pushpak Bhattacharyya. 2024.
\newblock \href {https://doi.org/10.18653/v1/2024.acl-long.228} {In-context mixing ({ICM}): Code-mixed prompts for multilingual {LLM}s}.
\newblock In \emph{Proceedings of the 62nd Annual Meeting of the Association for Computational Linguistics (Volume 1: Long Papers)}, pages 4162--4176, Bangkok, Thailand. Association for Computational Linguistics.

\bibitem[{Son et~al.(2024)Son, Lee, Kim, Kim, Lee, Yeom, Jung, Kim, and Kim}]{son-etal-2024-hae}
Guijin Son, Hanwool Lee, Suwan Kim, Huiseo Kim, Jae~cheol Lee, Je~Won Yeom, Jihyu Jung, Jung~woo Kim, and Songseong Kim. 2024.
\newblock \href {https://aclanthology.org/2024.lrec-main.704} {{HAE}-{RAE} bench: Evaluation of {K}orean knowledge in language models}.
\newblock In \emph{Proceedings of the 2024 Joint International Conference on Computational Linguistics, Language Resources and Evaluation (LREC-COLING 2024)}, pages 7993--8007, Torino, Italia. ELRA and ICCL.

\bibitem[{Srivastava and Singh(2021)}]{srivastava-singh-2021-challenges}
Vivek Srivastava and Mayank Singh. 2021.
\newblock \href {https://doi.org/10.18653/v1/2021.calcs-1.2} {Challenges and limitations with the metrics measuring the complexity of code-mixed text}.
\newblock In \emph{Proceedings of the Fifth Workshop on Computational Approaches to Linguistic Code-Switching}, pages 6--14, Online. Association for Computational Linguistics.

\bibitem[{Vivek et~al.(2024)Vivek, Ethayarajh, Yang, and Kiela}]{vivek-etal-2024-anchor}
Rajan Vivek, Kawin Ethayarajh, Diyi Yang, and Douwe Kiela. 2024.
\newblock \href {https://aclanthology.org/2024.eacl-long.95/} {Anchor points: Benchmarking models with much fewer examples}.
\newblock In \emph{Proceedings of the 18th Conference of the European Chapter of the Association for Computational Linguistics (Volume 1: Long Papers)}, pages 1576--1601, St. Julian{'}s, Malta. Association for Computational Linguistics.

\bibitem[{Wang et~al.(2025)Wang, Li, Zhou, Weng, Wang, Huang, Han, Feng, Deng, and Huang}]{wang-etal-2025-investigating-scaling}
Zhijun Wang, Jiahuan Li, Hao Zhou, Rongxiang Weng, Jingang Wang, Xin Huang, Xue Han, Junlan Feng, Chao Deng, and Shujian Huang. 2025.
\newblock \href {https://doi.org/10.18653/v1/2025.findings-acl.575} {Investigating and scaling up code-switching for multilingual language model pre-training}.
\newblock In \emph{Findings of the Association for Computational Linguistics: ACL 2025}, pages 11032--11046, Vienna, Austria. Association for Computational Linguistics.

\bibitem[{Wei et~al.(2023)Wei, Wang, Schuurmans, Bosma, Ichter, Xia, Chi, Le, and Zhou}]{cot}
Jason Wei, Xuezhi Wang, Dale Schuurmans, Maarten Bosma, Brian Ichter, Fei Xia, Ed~Chi, Quoc Le, and Denny Zhou. 2023.
\newblock \href {https://arxiv.org/abs/2201.11903} {Chain-of-thought prompting elicits reasoning in large language models}.
\newblock \emph{Preprint}, arXiv:2201.11903.

\bibitem[{Yong et~al.(2023)Yong, Zhang, Forde, Wang, Subramonian, Lovenia, Cahyawijaya, Winata, Sutawika, Cruz, Tan, Phan, Phan, Garcia, Solorio, and Aji}]{yong-etal-2023-prompting-code-mixed}
Zheng~Xin Yong, Ruochen Zhang, Jessica Forde, Skyler Wang, Arjun Subramonian, Holy Lovenia, Samuel Cahyawijaya, Genta Winata, Lintang Sutawika, Jan Christian~Blaise Cruz, Yin~Lin Tan, Long Phan, Long Phan, Rowena Garcia, Thamar Solorio, and Alham~Fikri Aji. 2023.
\newblock \href {https://aclanthology.org/2023.calcs-1.5} {Prompting multilingual large language models to generate code-mixed texts: The case of south {E}ast {A}sian languages}.
\newblock In \emph{Proceedings of the 6th Workshop on Computational Approaches to Linguistic Code-Switching}, pages 43--63, Singapore. Association for Computational Linguistics.

\bibitem[{Zhang et~al.(2023{\natexlab{a}})Zhang, Cahyawijaya, Cruz, Winata, and Aji}]{zhang-etal-2023-multilingual-notyet}
Ruochen Zhang, Samuel Cahyawijaya, Jan Christian~Blaise Cruz, Genta Winata, and Alham~Fikri Aji. 2023{\natexlab{a}}.
\newblock \href {https://doi.org/10.18653/v1/2023.emnlp-main.774} {Multilingual large language models are not (yet) code-switchers}.
\newblock In \emph{Proceedings of the 2023 Conference on Empirical Methods in Natural Language Processing}, pages 12567--12582, Singapore. Association for Computational Linguistics.

\bibitem[{Zhang et~al.(2023{\natexlab{b}})Zhang, Li, Cui, Cai, Liu, Fu, Huang, Zhao, Zhang, Chen et~al.}]{zhang2023siren}
Yue Zhang, Yafu Li, Leyang Cui, Deng Cai, Lemao Liu, Tingchen Fu, Xinting Huang, Enbo Zhao, Yu~Zhang, Yulong Chen, et~al. 2023{\natexlab{b}}.
\newblock Siren's song in the ai ocean: a survey on hallucination in large language models.
\newblock \emph{arXiv preprint arXiv:2309.01219}.

\bibitem[{Zhang et~al.(2024)Zhang, Lee, Fang, Yu, Jia, Jiang, and Barbieri}]{zhang-etal-2024-plug}
Zhihan Zhang, Dong-Ho Lee, Yuwei Fang, Wenhao Yu, Mengzhao Jia, Meng Jiang, and Francesco Barbieri. 2024.
\newblock \href {https://doi.org/10.18653/v1/2024.acl-long.379} {{PLUG}: Leveraging pivot language in cross-lingual instruction tuning}.
\newblock In \emph{Proceedings of the 62nd Annual Meeting of the Association for Computational Linguistics (Volume 1: Long Papers)}, pages 7025--7046, Bangkok, Thailand. Association for Computational Linguistics.

\bibitem[{Zheng et~al.(2023)Zheng, Chiang, Sheng, Zhuang, Wu, Zhuang, Lin, Li, Li, Xing, Zhang, Gonzalez, and Stoica}]{llm-as-judge}
Lianmin Zheng, Wei-Lin Chiang, Ying Sheng, Siyuan Zhuang, Zhanghao Wu, Yonghao Zhuang, Zi~Lin, Zhuohan Li, Dacheng Li, Eric.~P Xing, Hao Zhang, Joseph~E. Gonzalez, and Ion Stoica. 2023.
\newblock \href {https://arxiv.org/abs/2306.05685} {Judging llm-as-a-judge with mt-bench and chatbot arena}.
\newblock \emph{Preprint}, arXiv:2306.05685.

\end{thebibliography}
